\documentclass[10pt,twocolumn,letterpaper]{article}

\usepackage{cvpr,times,graphicx,amsmath,amssymb,multirow,array,xcolor,textcomp}
\usepackage[caption=false]{subfig}
\definecolor{citecolor}{RGB}{34,139,34}
\usepackage[pagebackref=false,breaklinks=true,letterpaper=true,colorlinks,
 citecolor=citecolor,bookmarks=false]{hyperref}
\usepackage[british,english,american]{babel}
\def\cvprPaperID{***}\cvprfinalcopy

\newcommand{\bd}[1]{\textbf{#1}}
\def\x{\times}
\def\B{\mathcal{B}}
\def\Xn{X^n}
\newcommand{\app}{\raise.17ex\hbox{$\scriptstyle\sim$}}
\newcommand{\symb}[1]{{\small\texttt{#1}}\xspace}
\newcolumntype{x}[1]{>{\centering\arraybackslash}p{#1pt}}
\newlength\savewidth\newcommand\shline{\noalign{\global\savewidth\arrayrulewidth
  \global\arrayrulewidth 1pt}\hline\noalign{\global\arrayrulewidth\savewidth}}
\newcommand{\tablestyle}[2]{\setlength{\tabcolsep}{#1}\renewcommand{\arraystretch}{#2}\centering\footnotesize}
\newcommand{\eqntight}[2]{\vspace{-1mm}\begin{equation}\label{eq:#1}#2\vspace{-1mm}\end{equation}\ignorespaces}
\makeatletter\renewcommand\paragraph{\@startsection{paragraph}{4}{\z@}
  {.5em \@plus1ex \@minus.2ex}{-.5em}{\normalfont\normalsize\bfseries}}\makeatother

\newcommand{\demph}[1]{\textcolor{gray}{#1}}

\begin{document}

\title{Accurate, Large Minibatch SGD:\\Training ImageNet in 1 Hour\vspace{-3mm}}

\author{
 Priya Goyal \qquad Piotr Doll\'ar \qquad Ross Girshick \qquad Pieter Noordhuis \\
 \quad Lukasz Wesolowski \quad Aapo Kyrola \quad Andrew Tulloch
 \quad Yangqing Jia \quad Kaiming He\vspace{4mm}\\
 Facebook\vspace{-1mm}
}

\maketitle

\begin{abstract}
Deep learning thrives with large neural networks and large datasets. However, larger networks and larger datasets result in longer training times that impede research and development progress. Distributed synchronous SGD offers a potential solution to this problem by dividing SGD minibatches over a pool of parallel workers. Yet to make this scheme efficient, the per-worker workload must be large, which implies nontrivial growth in the SGD minibatch size. In this paper, we empirically show that on the ImageNet dataset large minibatches cause optimization difficulties, but when these are addressed the trained networks exhibit good generalization. Specifically, we show no loss of accuracy when training with large minibatch sizes up to 8192 images. To achieve this result, we adopt a hyper-parameter-free linear scaling rule for adjusting learning rates as a function of minibatch size and develop a new warmup scheme that overcomes optimization challenges early in training. With these simple techniques, our Caffe2-based system trains ResNet-50 with a minibatch size of 8192 on 256 GPUs in one hour, while matching small minibatch accuracy. Using commodity hardware, our implementation achieves $\app$90\% scaling efficiency when moving from 8 to 256 GPUs. Our findings enable training visual recognition models on internet-scale data with high efficiency.
\end{abstract}

\section{Introduction}

Scale matters. We are in an unprecedented era in AI research history in which the increasing data and model scale is rapidly improving accuracy in computer vision \cite{Krizhevsky2012, Zeiler2014, Sermanet2014, Simonyan2015, Szegedy2015, He2016}, speech \cite{Hinton2012a, Xiong2016}, and natural language processing \cite{Collobert2011, Wu2016}. Take the profound impact in computer vision as an example: visual representations learned by deep convolutional neural networks \cite{LeCun1989, Krizhevsky2012} show excellent performance on previously challenging tasks like ImageNet classification \cite{Russakovsky2015} and can be transferred to difficult perception problems such as object detection and segmentation \cite{Donahue2014, Girshick2014, Long2015}. Moreover, this pattern generalizes: larger datasets and neural network architectures consistently yield improved accuracy across all tasks that benefit from pre-training \cite{Krizhevsky2012, Zeiler2014, Sermanet2014, Simonyan2015, Szegedy2015, He2016}. But as model and data scale grow, so does training time; discovering the potential and limits of large-scale deep learning requires developing novel techniques to keep training time manageable.

\begin{figure}[t]\centering
\includegraphics[width=.99\linewidth]{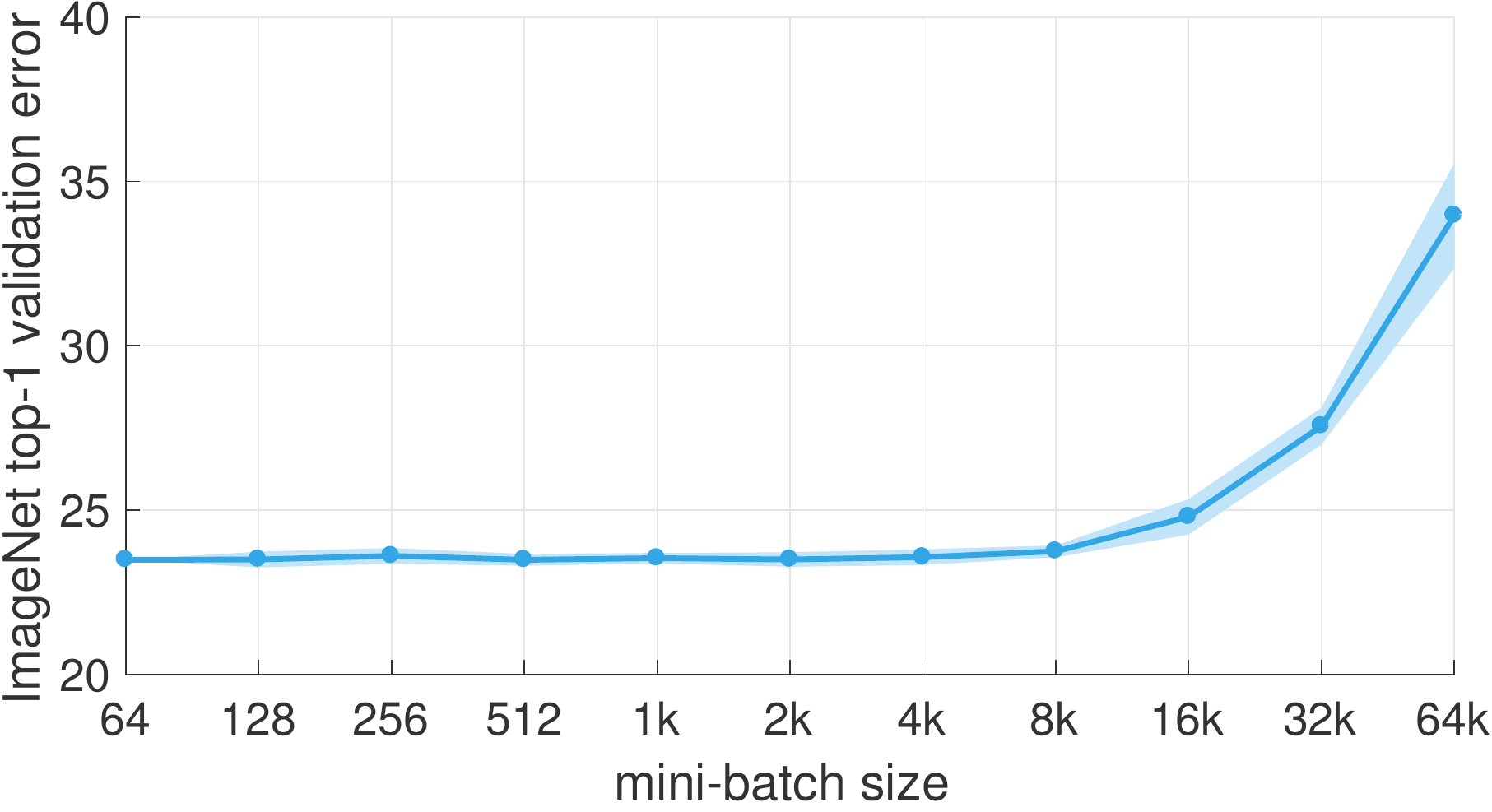}
\caption{\bd{ImageNet top-1 validation error \vs minibatch size.} Error range of plus/minus \emph{two} standard deviations is shown. We present a simple and general technique for scaling distributed synchronous SGD to minibatches of up to 8k images \emph{while maintaining the top-1 error of small minibatch training}. For all minibatch sizes we set the learning rate as a \emph{linear} function of the minibatch size and apply a simple warmup phase for the first few epochs of training. All other hyper-parameters are kept fixed. Using this simple approach, accuracy of our models is invariant to minibatch size (up to an 8k minibatch size). Our techniques enable a linear reduction in training time with \app90\% efficiency as we scale to large minibatch sizes, allowing us to train an accurate 8k minibatch ResNet-50 model in 1 hour on 256 GPUs.}
\label{fig:batchsize}
\end{figure}

The goal of this report is to demonstrate the feasibility of, and to communicate a practical guide to, large-scale training with distributed \emph{synchronous} stochastic gradient descent (SGD). As an example, we scale ResNet-50 \cite{He2016} training, originally performed with a minibatch size of 256 images (using 8 Tesla P100 GPUs, training time is 29 hours), to larger minibatches (see Figure~\ref{fig:batchsize}). In particular, we show that \emph{with a large minibatch size of 8192, we can train ResNet-50 in 1 hour using 256 GPUs while maintaining the same level of accuracy as the 256 minibatch baseline}. While distributed synchronous SGD is now commonplace, no existing results show that generalization accuracy can be maintained with minibatches as large as 8192 or that such high-accuracy models can be trained in such short time.

To tackle this unusually large minibatch size, we employ a simple and hyper-parameter-free \emph{linear scaling rule} to adjust the learning rate. While this guideline is found in earlier work \cite{Krizhevsky2014,Bottou2016}, its empirical limits are not well understood and informally we have found that it is not widely known to the research community. To successfully apply this rule, we present a new \emph{warmup} strategy, \ie, a strategy of using lower learning rates at the start of training \cite{He2016}, to overcome early optimization difficulties. Importantly, not only does our approach match the baseline \emph{validation} error, \emph{but also yields training error curves that closely match the small minibatch baseline.} Details are presented in \S\ref{sec:sgd}.

Our comprehensive experiments in \S\ref{sec:exp} show that \emph{optimization} difficulty is the main issue with large minibatches, rather than poor \emph{generalization} (at least on ImageNet), in contrast to some recent studies \cite{Keskar2017}. Additionally, we show that the linear scaling rule and warmup generalize to more complex tasks including object detection and instance segmentation \cite{Girshick2015,Ren2015,He2017,Long2015}, which we demonstrate via the recently developed Mask R-CNN \cite{He2017}. We note that a robust and successful guideline for addressing a wide range of minibatch sizes has not been presented in previous work.

While the strategy we deliver is simple, its successful application requires correct implementation with respect to seemingly minor and often not well understood implementation details within deep learning libraries. Subtleties in the implementation of SGD can lead to incorrect solutions that are difficult to discover. To provide more helpful guidance we describe common pitfalls and the relevant implementation details that can trigger these traps in \S\ref{sec:remarks}.

Our strategy applies regardless of framework, but achieving efficient linear scaling requires nontrivial communication algorithms. We use the open-source \emph{Caffe2}\footnote{\url{http://www.caffe2.ai}} deep learning framework and \emph{Big Basin} GPU servers~\cite{Lee2017}, which operates efficiently using standard Ethernet networking (as opposed to specialized network interfaces). We describe the systems algorithms that enable our approach to operate near its full potential in \S\ref{sec:commhw}.

The practical advances described in this report are helpful across a range of domains. In an industrial domain, our system unleashes the potential of training visual models from internet-scale data, enabling training with \emph{billions of images per day}. Of equal importance, in a research domain, we have found it to simplify migrating algorithms from a single-GPU to a multi-GPU implementation without requiring hyper-parameter search, \eg in our experience migrating Faster R-CNN \cite{Ren2015} and ResNets \cite{He2016} from 1 to 8 GPUs.

\section{Large Minibatch SGD}\label{sec:sgd}

We start by reviewing the formulation of Stochastic Gradient Descent (SGD), which will be the foundation of our discussions in the following sections. We consider supervised learning by minimizing a loss $L(w)$ of the form:
 \eqntight{loss}{ L(w) = \frac{1}{|X|}\sum_{x\in X} l(x,w). }
Here $w$ are the weights of a network, $X$ is a labeled training set, and $l(x,w)$ is the loss computed from samples $x\in X$ and their labels $y$. Typically $l$ is the sum of a classification loss (\eg, cross-entropy) and a regularization loss on $w$.

\emph{Minibatch Stochastic Gradient Descent} \cite{Robbins1951}, usually referred to as simply as SGD in recent literature even though it operates on minibatches, performs the following update:
 \eqntight{sgd}{ w_{t+1} = w_{t} - \eta \frac{1}{n}\sum_{x\in\B} \nabla l(x,w_t). }
Here $\B$ is a \emph{minibatch} sampled from $X$ and $n=|\B|$ is the \emph{minibatch size}, $\eta$ is the \emph{learning rate}, and $t$ is the iteration index. Note that in practice we use momentum SGD; we return to a discussion of momentum in \S\ref{sec:remarks}.

\subsection{Learning Rates for Large Minibatches}\label{sec:lr_rule}

Our goal is to use large minibatches in place of small minibatches while \emph{maintaining training and generalization accuracy}. This is of particular interest in distributed learning, because it can allow us to scale to multiple workers\footnote{We use the terms `worker' and `GPU' interchangeably in this work, although other implementations of a `worker' are possible. `Server' denotes a set of 8 GPUs that does not require communication over a network.} using simple data parallelism without reducing the per-worker workload and without sacrificing model accuracy.

As we will show in comprehensive experiments, we found that the following learning rate scaling rule is surprisingly effective for a broad range of minibatch sizes:
\begin{center}
\fbox{\begin{minipage}{.85\linewidth}\centering
 \emph{\textbf{Linear Scaling Rule:} When the minibatch size is multiplied by $k$, multiply the learning rate by $k$.}
\end{minipage}}
\end{center}
All other hyper-parameters (weight decay, \etc) are kept unchanged. As we will show in \S\ref{sec:exp}, the \emph{linear scaling rule} can help us to not only match the accuracy between using small and large minibatches, but equally importantly, to largely match their training curves, which enables rapid debugging and comparison of experiments prior to convergence.

\paragraph{Interpretation.} We present an informal discussion of the linear scaling rule and why it may be effective. Consider a network at iteration $t$ with weights $w_t$, and a sequence of $k$ minibatches $\B_j$ for $0\le j<k$ each of size $n$. We compare the effect of executing $k$ SGD iterations  with \emph{small minibatches} $\B_j$ and learning rate $\eta$ versus a single iteration with a \emph{large minibatch} $\cup_j \B_j$ of size $kn$ and learning rate $\hat\eta$.

According to (\ref{eq:sgd}), after $k$ iterations of SGD with learning rate $\eta$ and a minibatch size of $n$ we have:
 \eqntight{sgd_kxn}{ w_{t+k} = w_t - \eta \frac{1}{n} \sum_{j<k} \sum_{x\in\B_j} \nabla l(x,w_{t+j}). }
On the other hand, taking a single step with the large minibatch $\cup_j \B_j$ of size $kn$ and learning rate $\hat\eta$ yields:
 \eqntight{sgd_kn}{ \hat w_{t+1} = w_t -  \hat\eta \frac{1}{kn} \sum_{j<k} \sum_{x\in\B_j} \nabla l(x,w_t). }
As expected, the updates differ, and it is \emph{unlikely} that $\hat w_{t+1} = w_{t+k}$. However, if we \emph{could} assume $\nabla l(x,w_t) \approx \nabla l(x,w_{t+j})$ for $j<k$, then setting $\hat\eta=k\eta$ would yield $\hat w_{t+1} \approx w_{t+k}$, and the updates from small and large minibatch SGD would be similar. Although this is a strong assumption, we emphasize that if it were true the two updates are similar \emph{only} if we set $\hat\eta=k\eta$.

The above interpretation gives intuition for one case where we may hope the linear scaling rule to apply. In our experiments with $\hat \eta = k \eta$ (and warmup), small and large minibatch SGD not only result in models with the same final accuracy, but also, the training curves match closely. Our empirical results suggest that the above approximation \emph{might} be valid in large-scale, real-world data.

However, there are at least two cases when the condition $\nabla l(x,w_t) \approx \nabla l(x,w_{t+j})$ will clearly not hold. First, in initial training when the network is changing rapidly, it does not hold. We address this by using a warmup phase, discussed in \S\ref{sec:warmup}. Second, minibatch size cannot be scaled indefinitely: while results are stable for a large range of sizes, beyond a certain point accuracy degrades rapidly. Interestingly, this point is as large as $\app$8k in ImageNet experiments.

\paragraph{Discussion.} The above linear scaling rule was adopted by Krizhevsky \cite{Krizhevsky2014}, if not earlier. However, Krizhevsky reported a 1\% increase of error when increasing the minibatch size from 128 to 1024, whereas we show how to maintain accuracy across a much broader regime of minibatch sizes. Chen \etal  \cite{Chen2016} presented a comparison of numerous distributed SGD variants, and although their work also employed the linear scaling rule, it did not establish a small minibatch baseline. Li \cite{MuLiPhDThesis2017} (\S4.6) showed distributed ImageNet training with minibatches up to 5120 without a loss in accuracy after convergence. However, their work did not demonstrate a hyper-parameter search-free rule for adjusting the learning rate as a function of minibatch size, which is a central contribution of our work.

In recent work, Bottou \etal \cite{Bottou2016} (\S4.2) review theoretical tradeoffs of minibatching and show that with the linear scaling rule, solvers follow the same training curve as a function of number of examples seen, and suggest the learning rate should not exceed a maximum rate independent of minibatch size (which justifies warmup). Our work empirically tests these theories with unprecedented minibatch sizes.

\subsection{Warmup}\label{sec:warmup}

As we discussed, for large minibatches (\eg, 8k) the linear scaling rule breaks down when the network is changing rapidly, which commonly occurs in early stages of training. We find that this issue can be alleviated by a properly designed \emph{warmup} \cite{He2016}, namely, a strategy of using less aggressive learning rates at the start of training.

\paragraph{Constant warmup.} The warmup strategy presented in \cite{He2016} uses a low \emph{constant} learning rate for the first few epochs of training. As we will show in \S\ref{sec:exp}, we have found constant warmup particularly helpful for prototyping object detection and segmentation methods \cite{Girshick2015, Ren2015, Lin2017, He2017} that fine-tune pre-trained layers together with newly initialized layers.

In our ImageNet experiments with a large minibatch of size $kn$, we have tried to train with the low learning rate of $\eta$ for the first 5 epochs and then return to the target learning rate of $\hat\eta=k\eta$. However, given a large $k$, we find that this constant warmup is not sufficient to solve the optimization problem, and a transition out of the low learning rate warmup phase can cause the training error to spike. This leads us to propose the following gradual warmup.

\paragraph{Gradual warmup.} We present an alternative warmup that \emph{gradually} ramps up the learning rate from a small to a large value. This ramp avoids a sudden increase of the learning rate, allowing healthy convergence at the start of training. In practice, with a large minibatch of size $kn$, we start from a learning rate of $\eta$ and increment it by a constant amount at each iteration such that it reaches $\hat\eta=k\eta$ after 5 epochs (results are robust to the exact duration of warmup). After the warmup, we go back to the original learning rate schedule.

\subsection{Batch Normalization with Large Minibatches}\label{section:BN}

Batch Normalization (BN) \cite{Ioffe2015} computes statistics along the minibatch dimension: this breaks the independence of each sample's loss, and changes in minibatch size change the underlying definition of the loss function being optimized. In the following we will show that a commonly used `shortcut', which may appear to be a practical consideration to avoid communication overhead, is actually necessary for preserving the loss function when changing minibatch size.

We note that (\ref{eq:loss}) and (\ref{eq:sgd}) assume the per-sample loss $l(x,w)$ is independent of all other samples. This is \emph{not} the case when BN is performed and activations are computed across samples. We write $l_\B(x,w)$ to denote that the loss of a single sample $x$ depends on the statistics of all samples in its minibatch $\B$. We denote the loss over a single minibatch $\B$ of size $n$ as $L(\B,w) = \frac{1}{n}\sum_{x\in\B} l_\B(x,w)$. With BN, the training set can be thought of as containing all distinct subsets of size $n$ drawn from the original training set $X$, which we denote as $\Xn$. The training loss $L(w)$ then becomes:
 \eqntight{totallossbn}{ L(w) = \frac{1}{|\Xn|}\sum_{\B\in\Xn}L(\B,w). }
If we view $\B$ as a `single sample' in $\Xn$, then the loss of each single sample $\B$ is computed \emph{independently}.

Note that the minibatch size $n$ over which the BN statistics are computed is a key component of the loss: if the per-worker minibatch sample size $n$ is changed, \emph{it changes the underlying loss function $L$ that is optimized}. More specifically, the mean/variance statistics computed by BN with different $n$ exhibit different levels of random variation.

In the case of distributed (and multi-GPU) training, if the per-worker sample size $n$ is kept fixed and the total minibatch size is $kn$, it can be viewed a minibatch of $k$ samples with each sample $\B_j$ independently selected from $\Xn$, so the underlying loss function is unchanged and is still defined in $\Xn$.
Under this point of view, in the BN setting after seeing $k$ minibatches $\B_j$, (\ref{eq:sgd_kxn}) and (\ref{eq:sgd_kn}) become:
 \eqntight{sgdbn_k}{ w_{t+k} = w_{t} - \eta \sum_{j < k} \nabla L(\B_j,w_{t+j}), }
 \eqntight{sgdbn}{ \hat w_{t+1} = w_{t} - \hat \eta \frac{1}{k}\sum_{j<k} \nabla L(\B_j,w_t). }
Following similar logic as in \S\ref{sec:lr_rule}, we set $\hat \eta = k\eta$ and \emph{we keep the per-worker sample size $n$ constant when we change the number of workers $k$}.

In this work, we use $n=32$ which has performed well for a wide range of datasets and networks \cite{Ioffe2015,He2016}. If $n$ is adjusted, it should be viewed as a hyper-parameter of BN, not of distributed training. We also note that \emph{the BN statistics should \emph{not} be computed across all workers}, not only for the sake of reducing communication, but also for maintaining the same underlying loss function being optimized.

\section{Subtleties and Pitfalls of Distributed SGD}\label{sec:remarks}

In practice a distributed implementation has many subtleties. Many common implementation errors change the definitions of hyper-parameters, leading to models that train but whose error may be higher than expected, and such issues can be difficult to discover. While the remarks below are straightforward, they are important to consider explicitly to faithfully implement the underlying solver.

\paragraph{Weight decay.} Weight decay is actually the outcome of the gradient of an L2-regularization term \emph{in the loss function}. More formally, the per-sample loss in (\ref{eq:loss}) can be written as $l(x,w) = \frac{\lambda}{2} \| w\|^2 + \varepsilon(x,w)$. Here $\frac{\lambda}{2} \| w\|^2$ is the sample-independent L2 regularization on the weights and $\varepsilon(x,w)$ is a sample-dependent term such as the cross-entropy loss. The SGD update in (\ref{eq:sgd}) can be written as:
 \eqntight{sgd_wd}{ w_{t+1} = w_{t} - \eta\lambda w_t - \eta \frac{1}{n}\sum_{x\in\B} \nabla \varepsilon(x,w_t). }
In practice, usually only the sample-dependent term $\sum \nabla \varepsilon(x,w_t)$ is computed by backprop; the term $\lambda w_t$ is computed \emph{separately} and added to the aggregated gradients contributed by $\varepsilon(x,w_t)$. If there is no weight decay term, there are many equivalent ways of scaling the learning rate, including scaling the term $\varepsilon(x,w_t)$. However, as can be seen from (\ref{eq:sgd_wd}), in general this is \emph{not} the case. We summarize these observations in the following remark:
\begin{center}\begin{minipage}{.75\linewidth}\centering
 \emph{Remark 1: Scaling the cross-entropy loss is \emph{not} equivalent to scaling the learning rate.}
\end{minipage}\end{center}

\paragraph{Momentum correction.} Momentum SGD is a commonly adopted modification to the vanilla SGD in (\ref{eq:sgd}). A reference implementation of momentum SGD has the following form:
\begin{align}\label{eq:sgd_mmtu}\begin{split}
 u_{t+1} &= m u_{t} + \frac{1}{n}\sum_{x\in\B}\nabla l(x,w_t)\\[-.5em]
 w_{t+1} &= w_{t} - \eta u_{t+1}.
\end{split}\end{align}
Here $m$ is the momentum decay factor and $u$ is the update tensor. A popular variant absorbs the learning rate $\eta$ into the update tensor. Substituting $v_t$ for $\eta u_t$ in (\ref{eq:sgd_mmtu}) yields:
\begin{align}\label{eq:sgd_mmtv}\begin{split}
 v_{t+1} &= m v_{t} + \eta\frac{1}{n}\sum_{x\in\B}\nabla l(x,w_t)\\[-.5em]
 w_{t+1} &= w_{t} - v_{t+1}.
\end{split}\end{align}
For a fixed $\eta$, the two are equivalent. However, we note that while $u$ only depends on the gradients and is independent of $\eta$, $v$ is entangled with $\eta$. When $\eta$ changes, to maintain equivalence with the reference variant in (\ref{eq:sgd_mmtu}), the update for $v$ should be: $v_{t+1} = m \frac{\eta_{t+1}}{\eta_{t}} v_{t} + \eta_{t+1} \frac{1}{n}\sum\nabla l(x,w_t)$. We refer to the factor $\frac{\eta_{t+1}}{\eta_{t}}$ as the \emph{momentum correction}. We found that this is especially important for stabilizing training when $\eta_{t+1}\gg\eta_t$, otherwise the history term $v_{t}$ is too small which leads to instability (for $\eta_{t+1}<\eta_t$ momentum correction is less critical). This leads to our second remark:
\begin{center}\begin{minipage}{.75\linewidth}\centering
 \emph{Remark 2: Apply momentum correction after changing learning rate if using (\ref{eq:sgd_mmtv}).}
\end{minipage}\end{center}

\paragraph{Gradient aggregation.} For $k$ workers each with a per-worker minibatch of size $n$, following (\ref{eq:sgd_kn}), gradient aggregation must be performed over the entire set of $kn$ examples according to $\frac{1}{kn}\sum_j\sum_{x\in\B_j} l(x,w_t)$. Loss layers are typically implemented to compute an average loss over their \emph{local} input, which amounts to computing a per-worker loss of $\sum l(x,w_t)/n$. Given this, correct aggregation requires \emph{averaging} the $k$ gradients in order to recover the missing $1/k$ factor. However, standard communication primitives like allreduce \cite{gropp1999} perform summing, not averaging. Therefore, it is more efficient to absorb the $1/k$ scaling into the loss, in which case only the loss's gradient with respect to its input needs to be scaled, removing the need to scale the entire gradient vector. We summarize this as follows:
\begin{center}\begin{minipage}{.8\linewidth}\centering
 \emph{Remark 3: Normalize the per-worker loss by \emph{total} minibatch size $kn$, \emph{not} per-worker size $n$.}
\end{minipage}\end{center}
We also note that it may be incorrect to `cancel $k$' by setting $\hat \eta = \eta$ (not $k\eta$) and normalizing the loss by $1/n$ (not $1/kn$), which can lead to incorrect weight decay (see \emph{Remark 1}).

\paragraph{Data shuffling.} SGD is typically analyzed as a process that samples data randomly \emph{with replacement}. In practice, common SGD implementations apply \emph{random shuffling} of the training set during each SGD epoch, which can give better results \cite{bottou2009, gurbuzbalaban2015}. To provide fair comparisons with baselines that use shuffling (\eg, \cite{He2016}), we ensure the samples in one epoch done by $k$ workers are from a single consistent random shuffling of the training set. To achieve this, for each epoch we use a random shuffling that is partitioned into $k$ parts, each of which is processed by one of the $k$ workers. Failing to correctly implement random shuffling in multiple workers may lead to noticeably different behavior, which may contaminate results and conclusions. In summary:
\begin{center}\begin{minipage}{.94\linewidth}\centering
 \emph{Remark 4: Use a single random shuffling of the training data (per epoch) that is divided amongst all $k$ workers.}
\end{minipage}\end{center}

\section{Communication}\label{sec:commhw}

In order to scale beyond the 8 GPUs in a single Big Basin server \cite{Lee2017}, gradient aggregation has to span across servers on a network. To allow for near perfect linear scaling, the aggregation must be performed \emph{in parallel} with backprop. This is possible because there is no data dependency between gradients across layers. Therefore, as soon as the gradient for a layer is computed, it is aggregated across workers, while gradient computation for the next layer continues (as discussed in \cite{Chen2016}). We give full details next.

\subsection{Gradient Aggregation}

For every gradient, aggregation is done using an \emph{allreduce} operation (similar to the MPI collective operation \emph{MPI\_Allreduce} \cite{gropp1999}). Before allreduce starts every GPU has its locally computed gradients and after allreduce completes every GPU has the sum of all $k$ gradients. As the number of parameters grows and compute performance of GPUs increases, it becomes harder to hide the cost of aggregation in the backprop phase. Training techniques to overcome these effects are beyond the scope of this work (\eg, quantized gradients \cite{Hubara2016}, Block-Momentum SGD \cite{ChenHuo2016}). However, at the scale of this work, collective communication was not a bottleneck, as we were able to achieve near-linear SGD scaling by using an optimized allreduce implementation.

Our implementation of allreduce consists of three phases for communication within and across servers: (1) buffers from the 8 GPUs within a server are summed into a single buffer for each server, (2) the results buffers are shared and summed across all servers, and finally (3) the results are broadcast onto each GPU. For the local reduction and broadcast in phases (1) and (3) we used NVIDIA Collective Communication Library (NCCL)\footnote{\url{https://developer.nvidia.com/nccl}} for buffers of size 256 KB or more and a simple implementation consisting of a number of GPU-to-host memory copies and a CPU reduction otherwise. NCCL uses GPU kernels to accelerate intraserver collectives, so this approach dedicates more time on the GPU to backprop while using the CPU resources that would otherwise have been idle to improve throughput.

For interserver allreduce, we implemented two of the best algorithms for bandwidth-limited scenarios: the \emph{recursive halving and doubling algorithm} \cite{rabenseifner2004optimization, thakur2005optimization} and the \emph{bucket algorithm} (also known as the ring algorithm) \cite{barnett1994interprocessor}. For both, each server sends and receives $2 \frac{p - 1}{p} b$ bytes of data, where $b$ is the buffer size in bytes and $p$ is the number of servers. While the halving/doubling algorithm consists of $2 \log_2(p)$ communication steps, the ring algorithm consists of $2 (p - 1)$ steps. This generally makes the halving/doubling algorithm faster in latency-limited scenarios (\ie, for small buffer sizes and/or large server counts). In practice, we found the halving/doubling algorithm to perform much better than the ring algorithm for buffer sizes up to a million elements (and even higher on large server counts). On 32 servers (256 GPUs), using halving/doubling led to a speedup of 3$\x$ over the ring algorithm.

The halving/doubling algorithm consists of a reduce-scatter collective followed by an allgather. In the first step of reduce-scatter, servers communicate in pairs (rank 0 with 1, 2 with 3, \etc), sending and receiving for different halves of their input buffers. For example, rank 0 sends the second half of its buffer to 1 and receives the first half of the buffer from 1. A reduction over the received data is performed before proceeding to the next step, where the distance to the destination rank is doubled while the data sent and received is halved. After the reduce-scatter phase is finished, each server has a portion of the final reduced vector.

This is followed by the allgather phase, which retraces the communication pattern from the reduce-scatter in reverse, this time simply concatenating portions of the final reduced vector. At each server, the portion of the buffer that was being sent in the reduce-scatter is received in the allgather, and the portion that was being received is now sent.

To support non-power-of-two number of servers, we used the \emph{binary blocks algorithm} \cite{rabenseifner2004optimization}. This is a generalized version of the halving/doubling algorithm where servers are partitioned into power-of-two blocks and two additional communication steps are used, one immediately after the intrablock reduce-scatter and one before the intrablock allgather.  Non-power-of-two cases have some degree of load imbalance compared to power-of-two, though in our runs we did not see significant performance degradation.

\subsection{Software}

The allreduce algorithms described are implemented in \emph{Gloo}\footnote{\url{https://github.com/facebookincubator/gloo}}, a library for collective communication. It supports multiple communication contexts, which means no additional synchronization is needed to execute multiple allreduce instances in parallel. Local reduction and broadcast (described as phases (1) and (3)) are pipelined with interserver allreduce where possible.

\emph{Caffe2} supports multi-threaded execution of the compute graph that represents a training iteration. Whenever there is no data dependency between subgraphs, multiple threads can execute those subgraphs in parallel. Applying this to backprop, local gradients can be computed in sequence, without dealing with allreduce or weight updates. This means that during backprop, the set of \emph{runnable subgraphs} may grow faster than we can execute them. For subgraphs that contain an allreduce run, all servers must choose to execute the same subgraph from the set of runnable subgraphs. Otherwise, we risk distributed deadlock where servers are attempting to execute non-intersecting sets of subgraphs. With allreduce being a \emph{collective} operation, servers would time out waiting. To ensure correct execution we impose a partial order on these subgraphs. This is implemented using a cyclical control input, where completion of the $n$-th allreduce unblocks execution of the $(n+c)$-th allreduce, with $c$ being the maximum number of concurrent allreduce runs. Note that this number should be chosen to be lower than the number of threads used to execute the full compute graph.

\subsection{Hardware}

We used Facebook's Big Basin \cite{Lee2017} GPU servers for our experiments. Each server contains 8 NVIDIA Tesla P100 GPUs that are interconnected with NVIDIA NVLink. For local storage, each server has 3.2TB of NVMe SSDs. For network connectivity, the servers have a Mellanox ConnectX-4 50Gbit Ethernet network card and are connected to Wedge100 \cite{Bagga2016} Ethernet switches.

We have found 50Gbit of network bandwidth sufficient for distributed synchronous SGD for ResNet-50, per the following analysis. ResNet-50 has approximately 25 million parameters. This means the total size of parameters is $25 \cdot 10^6 \cdot \text{sizeof(float)}=100\text{MB}$. Backprop for ResNet-50 on a single NVIDIA Tesla P100 GPU takes 120 ms. Given that allreduce requires \app2$\x$ bytes on the network compared to the value it operates on, this leads to a peak bandwidth requirement of $200\text{MB}/0.125\text{s}=1600\text{MB/s}$, or 12.8 Gbit/s, not taking into account communication overhead. When we add a smudge factor for network overhead, we reach a peak bandwidth requirement for ResNet-50 of \app15 Gbit/s.

As this peak bandwidth requirement only holds during backprop, the network is free to be used for different tasks that are less latency sensitive then aggregation (\eg reading data or saving network snapshots) during the forward pass.

\section{Main Results and Analysis}\label{sec:exp}

Our main result is that we can train ResNet-50 \cite{He2016} on ImageNet \cite{Russakovsky2015} using 256 workers in one hour, while matching the accuracy of small minibatch training. Applying the linear scaling rule along with a warmup strategy allows us to seamlessly scale between small and large minibatches (up to 8k images) without tuning additional hyper-parameters or impacting accuracy. In the following subsections we: (1) describe experimental settings, (2) establish the effectiveness of large minibatch training, (3) perform a deeper experimental analysis, (4) show our findings generalize to object detection/segmentation, and (5) provide timings.

\subsection{Experimental Settings}\label{sec:exp:settings}

The 1000-way ImageNet classification task \cite{Russakovsky2015} serves as our main experimental benchmark. Models are trained on the \app1.28 million training images and evaluated by top-1 error on the 50,000 validation images.

We use the ResNet-50 \cite{He2016} variant from \cite{Gross2016}, noting that the stride-2 convolutions are on 3$\x$3 layers instead of on 1$\x$1 layers as in \cite{He2016}. We use Nesterov momentum \cite{Nesterov2004} with $m$ of 0.9 following \cite{Gross2016} but note that standard momentum as was used in \cite{He2016} is equally effective. We use a weight decay $\lambda$ of 0.0001 and following \cite{He2016} we do not apply weight decay on the learnable BN coefficients (namely, $\gamma$ and $\beta$ in \cite{Ioffe2015}). In order to keep the training objective fixed, which depends on the BN batch size $n$ as described in \S\ref{section:BN}, we use $n = 32$ throughout, regardless of the overall minibatch size. As in \cite{Gross2016}, we compute the BN statistics using running average (with momentum 0.9).

All models are trained for 90 epochs regardless of minibatch sizes. We apply the \emph{linear scaling rule} from \S\ref{sec:lr_rule} and use a learning rate of $\eta=0.1\cdot\frac{kn}{256}$ that is linear in the minibatch size $kn$. With $k=8$ workers (GPUs) and $n=32$ samples per worker, $\eta=0.1$ as in \cite{He2016}. We call this number ($0.1\cdot\frac{kn}{256}$) the \emph{reference learning rate}, and reduce it by $1/10$ at the 30-th, 60-th, and 80-th epoch, similar to \cite{He2016}.

We adopt the initialization of \cite{He2015} for all convolutional layers. The 1000-way fully-connected layer is initialized by drawing weights from a zero-mean Gaussian with standard deviation of 0.01. We have found that although SGD with a small minibatch is not sensitive to initialization due to BN, this is not the case for a substantially large minibatch. Additionally we require an appropriate warmup strategy to avoid optimization difficulties in early training.

For BN layers, the learnable scaling coefficient $\gamma$ is initialized to be 1, \emph{except for each residual block's last BN where $\gamma$ is initialized to be 0.} Setting $\gamma=0$ in the last BN of each residual block causes the forward/backward signal initially to propagate through the identity shortcut of ResNets, which we found to ease optimization at the start of training. This initialization improves all models but is particularly helpful for large minibatch training as we will show.

We use scale and aspect ratio data augmentation \cite{Szegedy2015} as in \cite{Gross2016}. The network input image is a 224$\x$224 pixel random crop from an augmented image or its horizontal flip. The input image is normalized by the per-color mean and  standard deviation, as in \cite{Gross2016}.

\paragraph{Handling random variation.} As models are subject to random variation in training, we compute a model's error rate as the \emph{median} error of the final 5 epochs. Moreover, we report the mean and standard deviation (std) of the error from \emph{5 independent runs}. This gives us more confidence in our results and also provides a measure of model stability.

The random variation of ImageNet models has generally not been reported in previous work (largely due to resource limitations). We emphasize that ignoring random variation may cause unreliable conclusions, especially if results are from a single trial, or the best of many.

\paragraph{Baseline.} Under these settings, we establish a ResNet-50 baseline using $k=8$ (8 GPUs in one server) and $n=32$ images per worker (minibatch size of $kn=256$), as in \cite{He2016}. Our baseline has a top-1 validation error of 23.60\% $\pm$0.12. As a reference, ResNet-50 from \symb{fb.resnet.torch} \cite{Gross2016} has 24.01\% error, and that of the original ResNet paper \cite{He2016} has 24.7\% under weaker data augmentation.

\subsection{Optimization or Generalization Issues?}\label{sec:exp:optimization}
We establish our main results on large minibatch training by exploring optimization and generalization behaviors. We will demonstrate that with a proper warmup strategy, large minibatch SGD can both match the \emph{training curves} of small minibatch SGD and also match the \emph{validation} error. In other words, in our experiments both \emph{optimization} and \emph{generalization} of large minibatch training matches that of small minibatch training. Moreover, in \S\ref{sec:exp:detection} we will show that these models exhibit good generalization behavior to the object detection/segmentation transfer tasks, matching the transfer quality of small minibatch models.

For the following results, we use $k=256$ and $n=32$, which results in a minibatch size $kn=8$k (we use `1k' to denote 1024). As discussed, our baseline has a minibatch size of $kn=256$ and a reference learning rate of $\eta=0.1$. Applying the linear scaling rule gives $\eta=3.2$ as the reference learning rate for our large minibatch runs. We test three warmup strategies as discussed in \S\ref{sec:warmup}: \emph{no warmup}, \emph{constant warmup} with $\eta=0.1$ for 5 epochs, and \emph{gradual warmup} which starts with $\eta=0.1$ and is linearly increased to $\eta=3.2$ over 5 epochs. All models are trained from scratch and all other hyper-parameters are kept fixed. We emphasize that while better results for any particular minibatch size could be obtained by optimizing hyper-parameters for that case; \emph{our goal is to match errors across minibatch sizes by using a general strategy that avoids hyper-parameter tuning for each minibatch size}.

\begin{figure*}[t]\centering
 \subfloat[no warmup\label{fig:warmup:none}]{
  \includegraphics[width=0.358\textwidth]{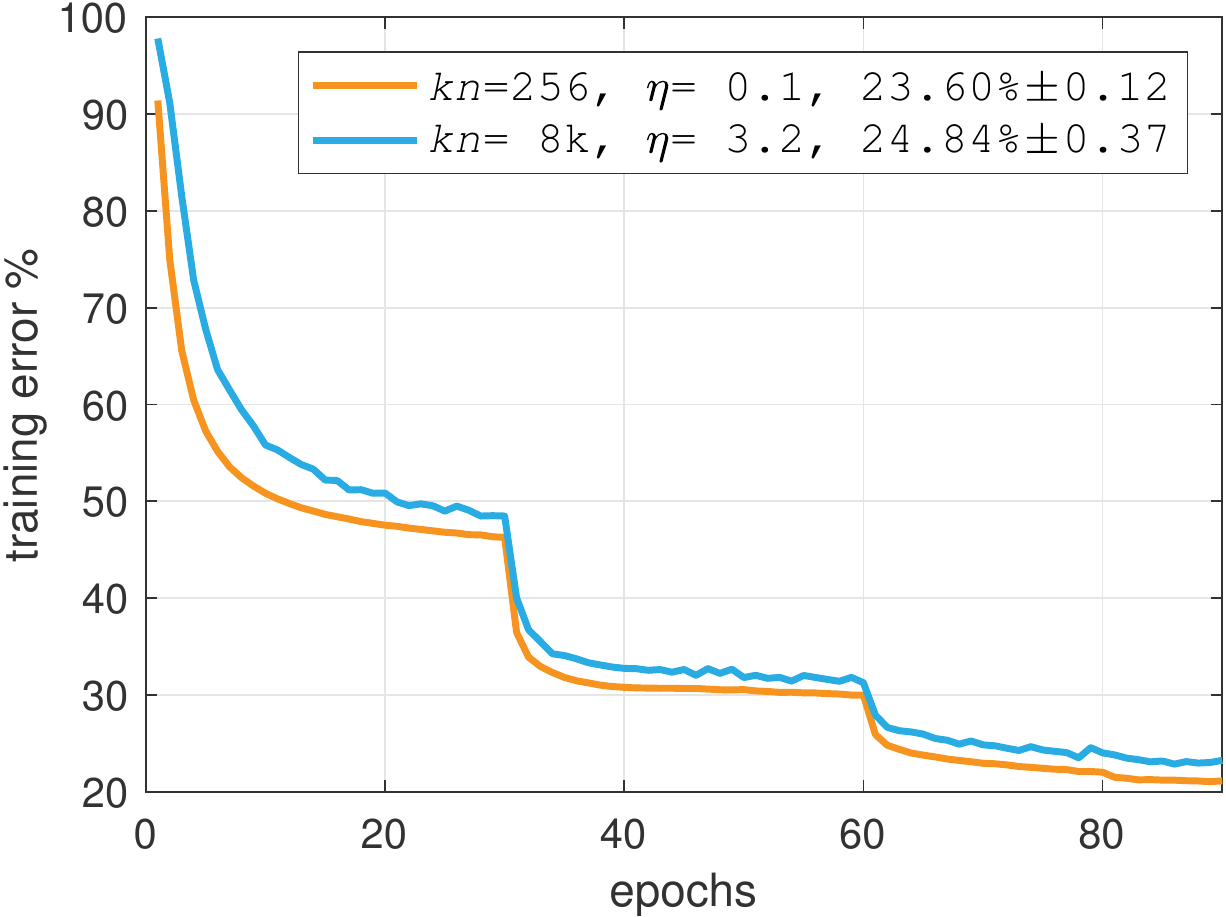}~}
 \subfloat[constant warmup\label{fig:warmup:constant}]{
  \includegraphics[width=0.32\textwidth]{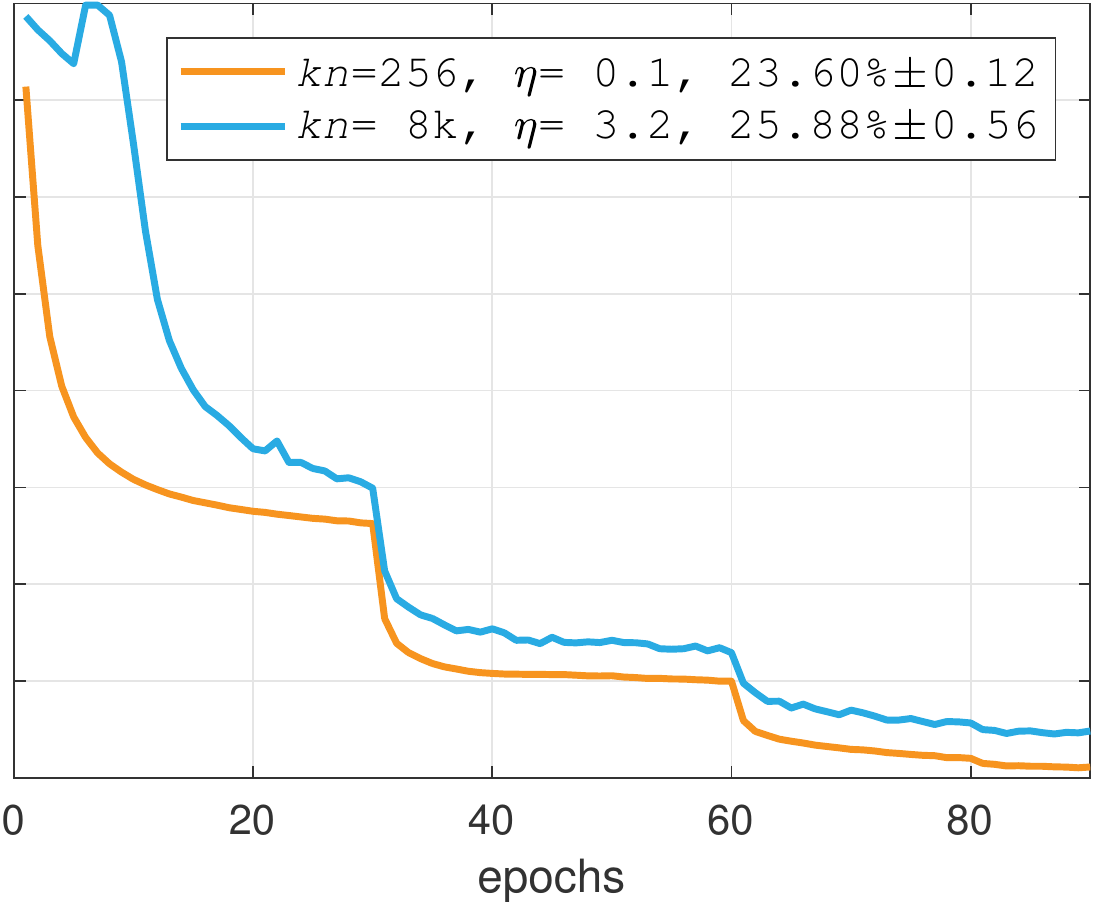}~}
 \subfloat[gradual warmup\label{fig:warmup:gradual}]{
  \includegraphics[width=0.32\textwidth]{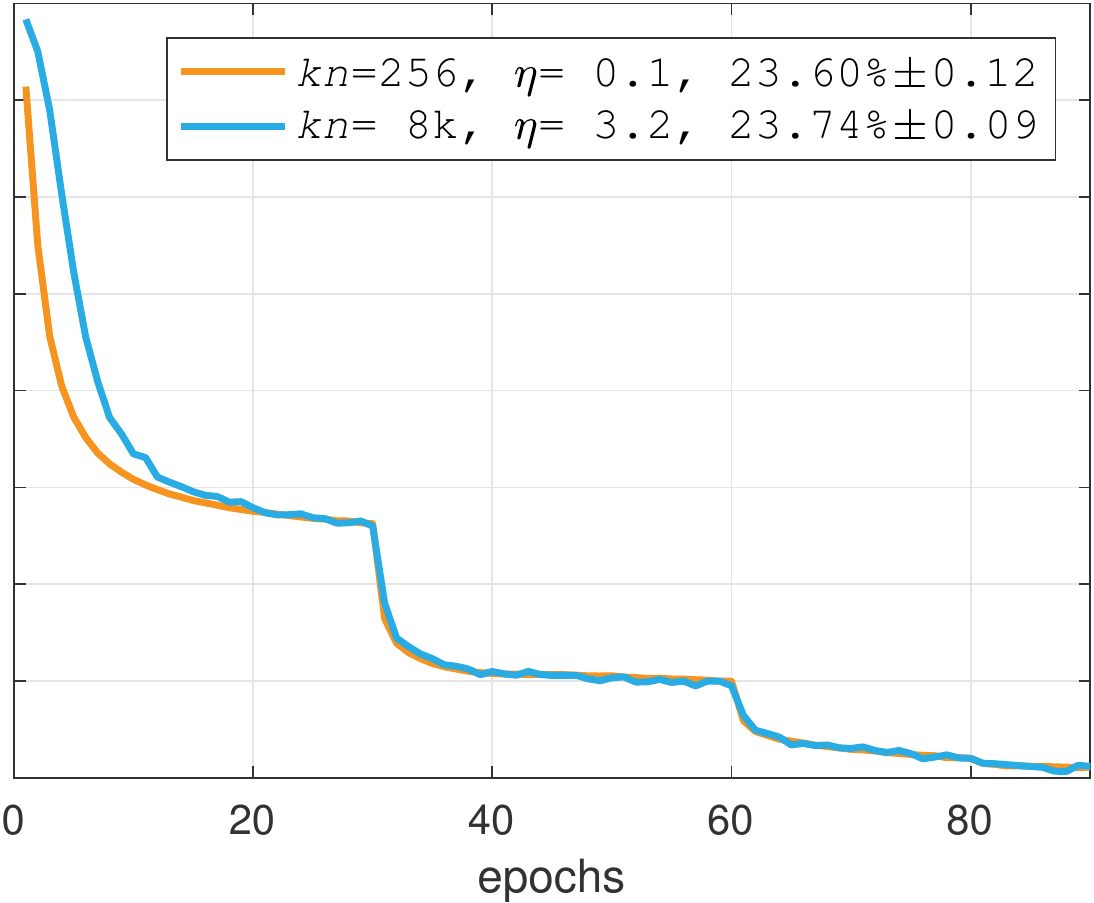}}\vspace{3mm}
\caption{\bd{Warmup.} Training error curves for minibatch size 8192 using various warmup strategies compared to minibatch size 256. \emph{Validation} error (mean$\pm$std of 5 runs) is shown in the legend, along with minibatch size $kn$ and reference learning rate $\eta$.}
\label{fig:warmup}\vspace{3mm}
\end{figure*}

\paragraph{Training error.} Training curves are shown in Figure~\ref{fig:warmup}. With no warmup (\ref{fig:warmup:none}), the training curve for large minibatch of $kn=8$k is inferior to training with a small minibatch of $kn=256$ across all epochs. A constant warmup strategy (\ref{fig:warmup:constant}) actually degrades results: although the small constant learning rate can decrease error during warmup, the error spikes immediately after and training never fully recovers.

Our main result is that with gradual warmup, large minibatch training error matches the baseline training curve obtained with small minibatches, see Figure~\ref{fig:warmup:gradual}. Although the large minibatch curve starts higher due to the low $\eta$ in the warmup phase, it catches up shortly thereafter. After about 20 epochs, the small and large minibatch training curves match closely. The comparison between no warmup and gradual warmup suggests that \emph{large minibatch sizes are challenged by optimization difficulties in early training} and if these difficulties are addressed, the training error and its curve can match a small minibatch baseline closely.

\begin{table}[t]
\tablestyle{5pt}{1.2}
\begin{tabular}{l|ccccc}
 & $k$ & $n$ & $kn$ & $\eta$ & top-1 error (\%)\\ [.1em]
\shline
 baseline (single server) & 8 & 32 & 256 & 0.1 & 23.60 $\pm$0.12 \\
\hline
no warmup, Figure~\ref{fig:warmup:none} & 256 & 32 & 8k & 3.2 & 24.84 $\pm$0.37 \\
constant warmup, Figure~\ref{fig:warmup:constant} & 256 & 32 & 8k & 3.2 & 25.88 $\pm$0.56 \\
gradual warmup, Figure~\ref{fig:warmup:gradual} & 256 & 32 & 8k & 3.2 & 23.74 $\pm$0.09 \\
\end{tabular}\vspace{3mm}
\caption{\bd{Validation error on ImageNet using ResNet-50} (mean and std computed over 5 trials). We compare the small minibatch model ($kn$=256) with large minibatch models ($kn$=8k) with various warmup strategies. Observe that the top-1 validation error for small and large minibatch training (with gradual warmup) is quite close: 23.60\% $\pm$0.12 \vs 23.74\% $\pm$0.09, respectively.}\label{tab:val}
\end{table}

\paragraph{Validation error.} Table~\ref{tab:val} shows the \emph{validation error} for the three warmup strategies. The no-warmup variant has $\app$1.2\% higher validation error than the baseline which is likely caused by the $\app2.1\%$ increase in training error (Figure~\ref{fig:warmup:none}), rather than overfitting or other causes for poor generalization. This argument is further supported by our gradual warmup experiment. The gradual warmup variant has a \emph{validation} error within 0.14\% of the baseline (noting that std of these estimates is \app0.1\%). Given that the final training errors (Figure~\ref{fig:warmup:gradual}) match nicely in this case, it shows that \emph{if the optimization issues are addressed, there is no apparent generalization degradation observed using large minibatch training}, even if the minibatch size goes from 256 to 8k.

\begin{figure*}[t]\centering
  \includegraphics[width=0.358\textwidth]{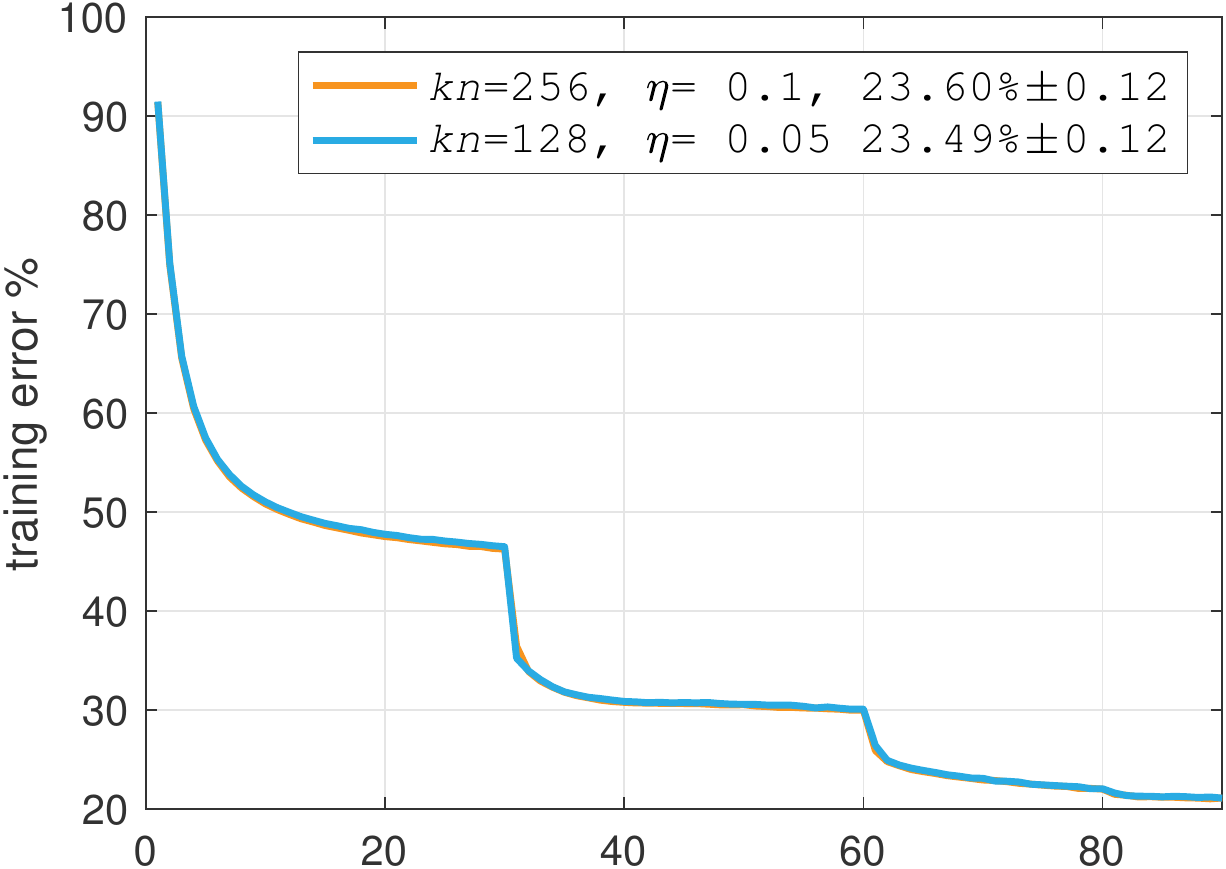}~
  \includegraphics[width=0.32\textwidth]{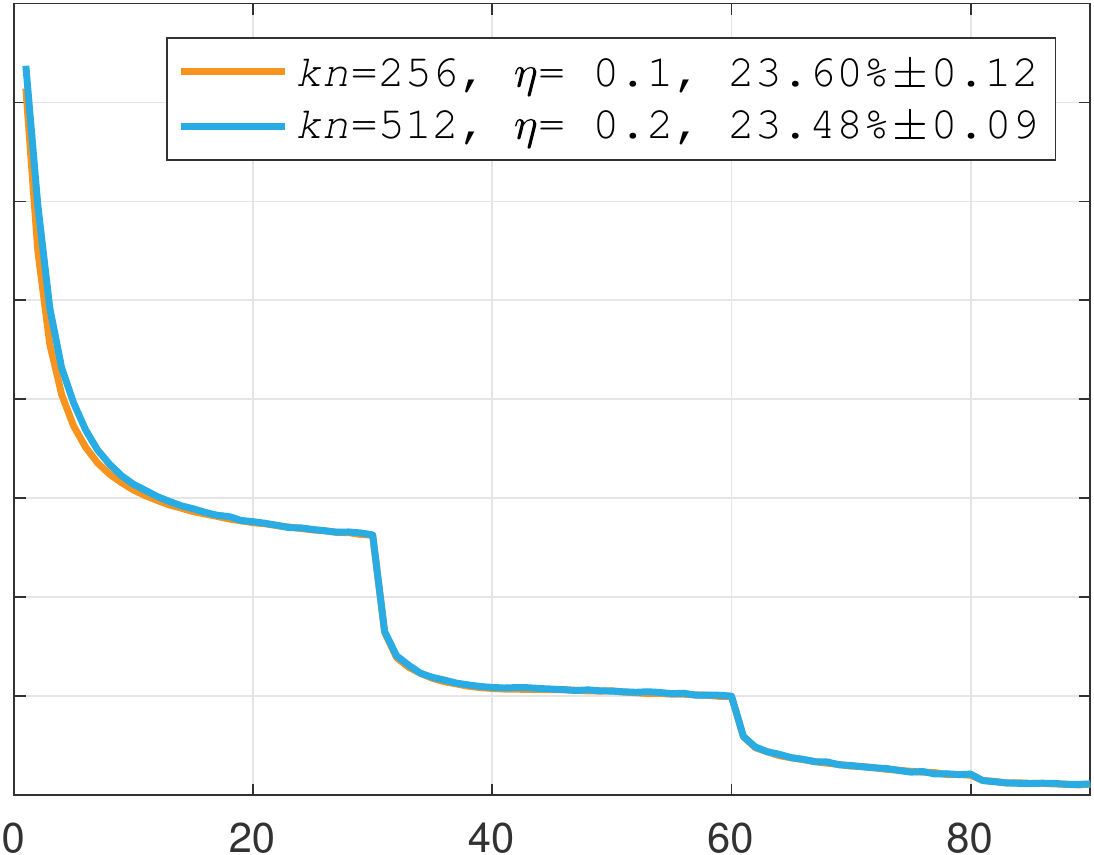}~
  \includegraphics[width=0.32\textwidth]{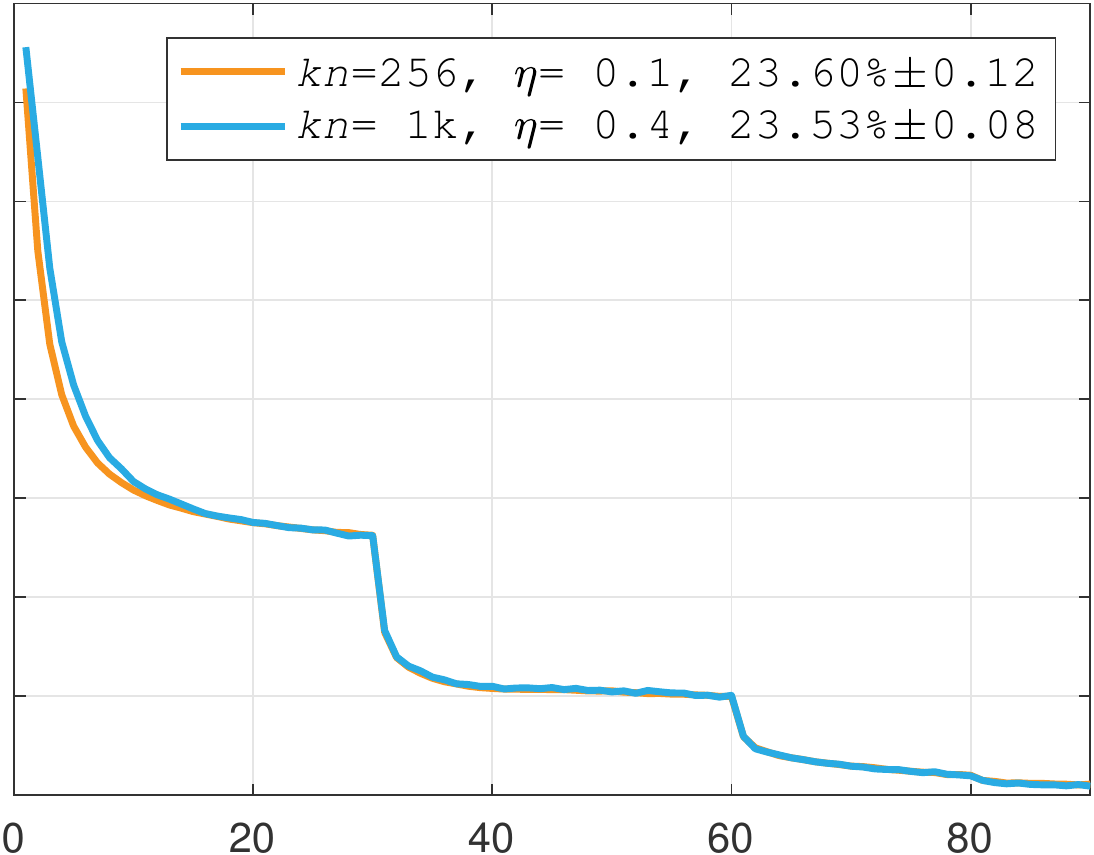}\\
  \includegraphics[width=0.358\textwidth]{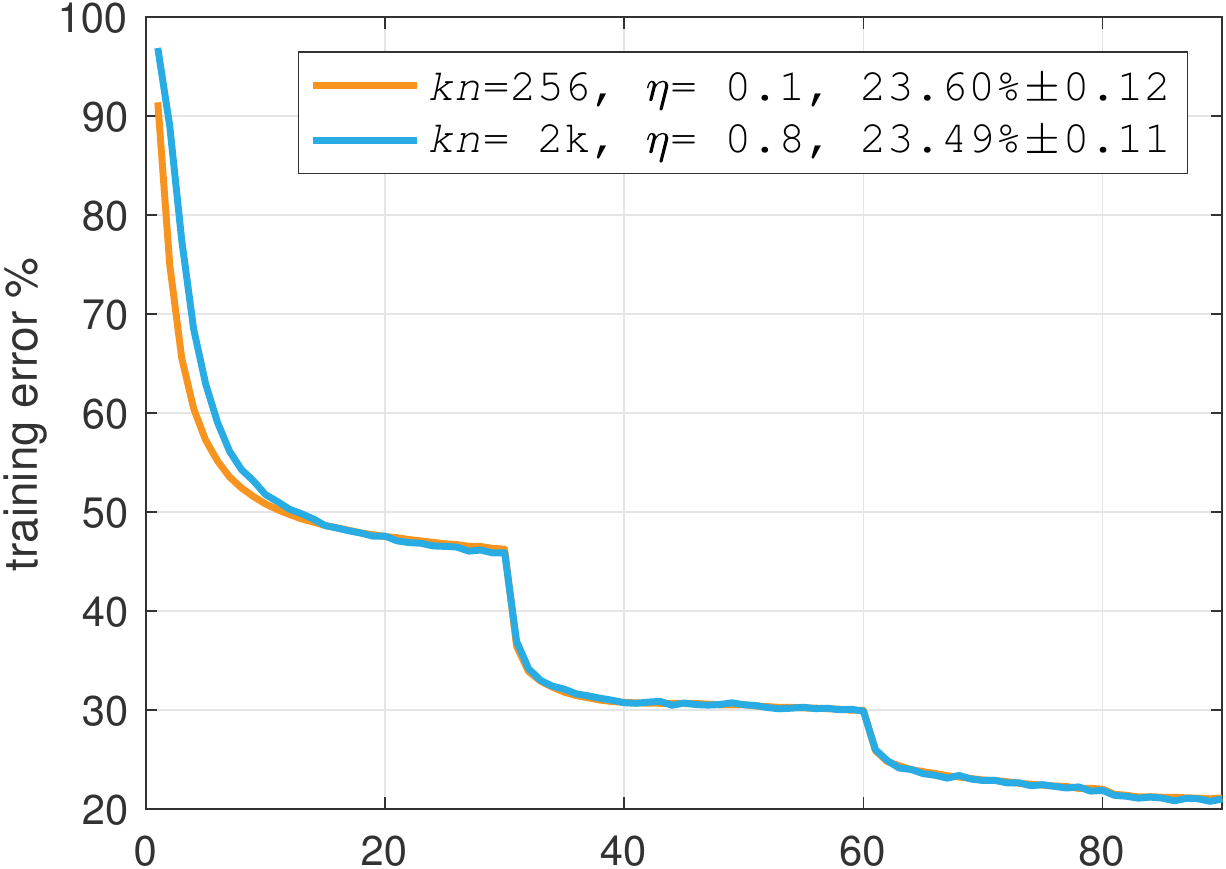}~
  \includegraphics[width=0.32\textwidth]{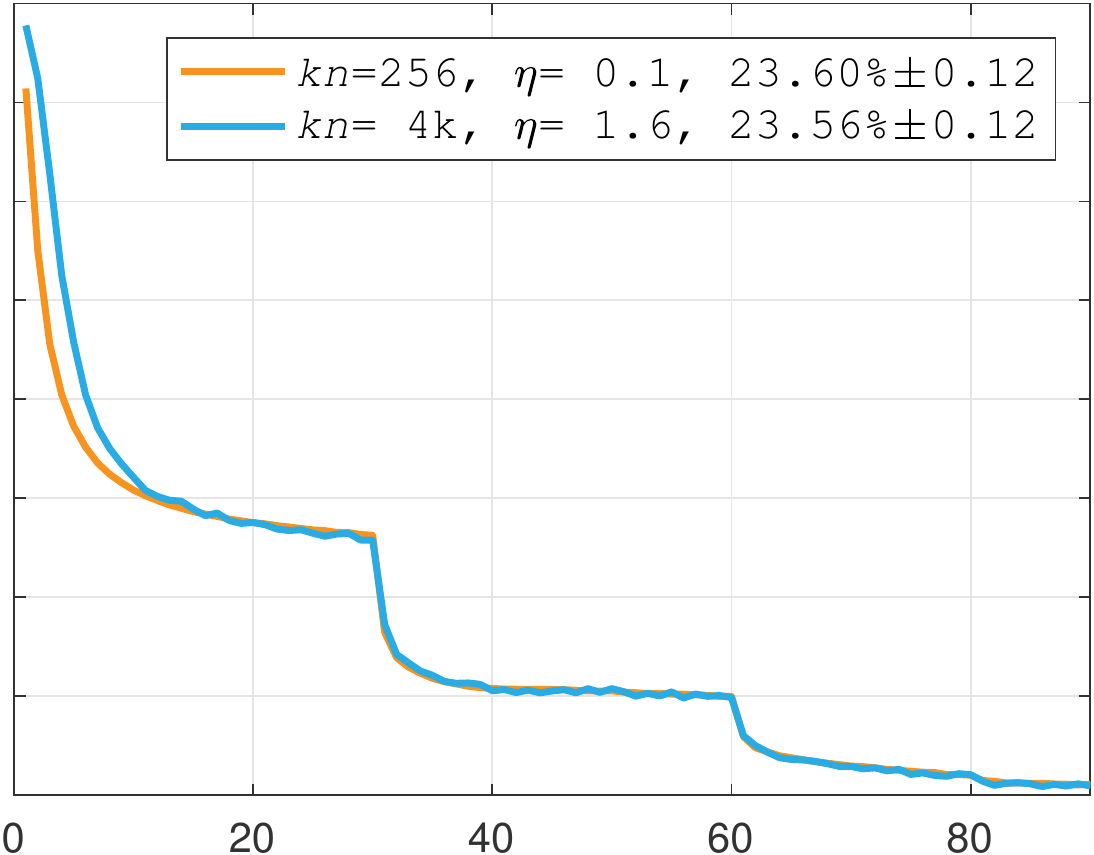}~
  \includegraphics[width=0.32\textwidth]{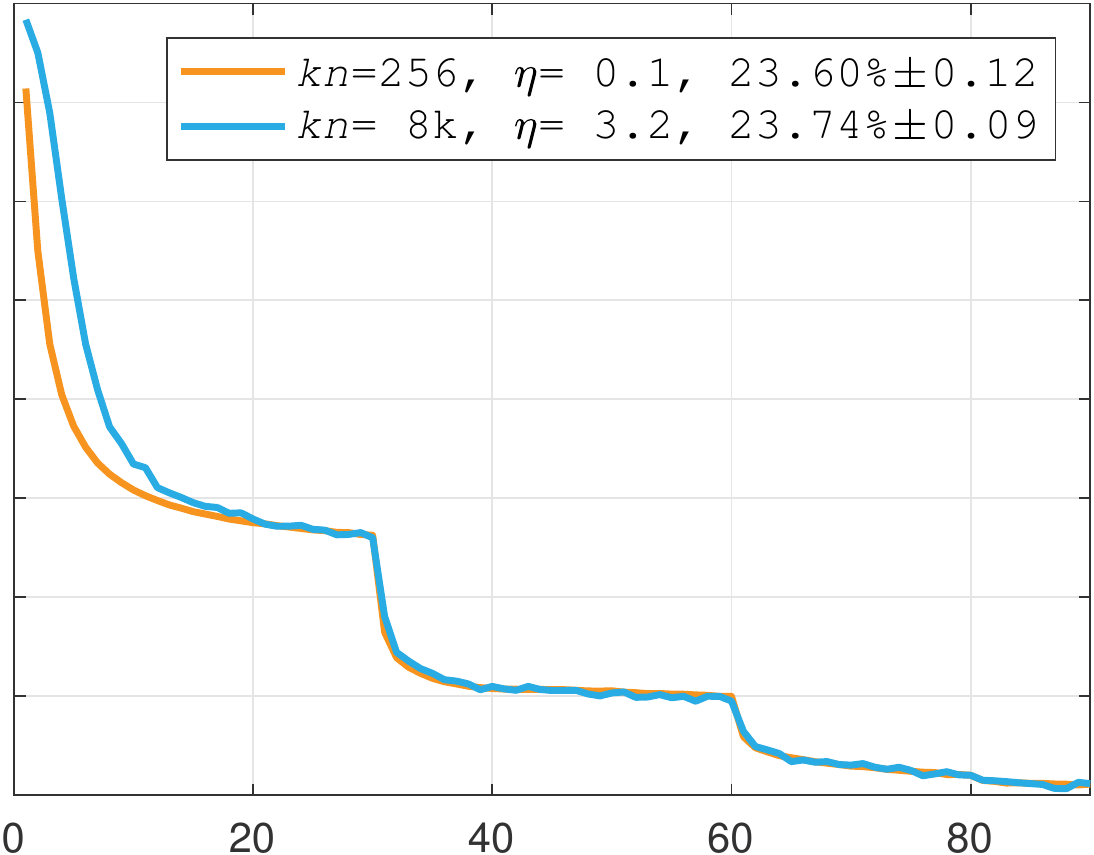}\\
  \includegraphics[width=0.358\textwidth]{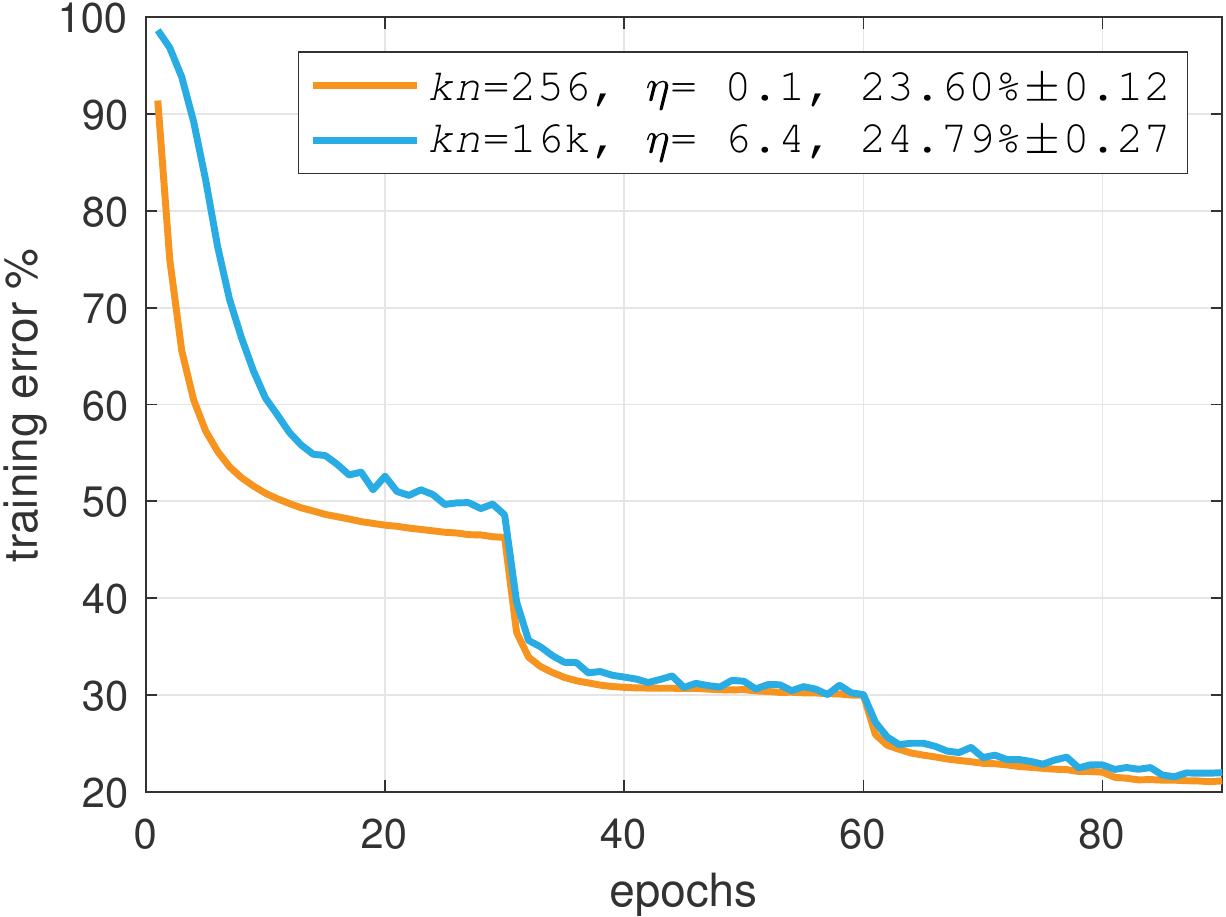}~
  \includegraphics[width=0.32\textwidth]{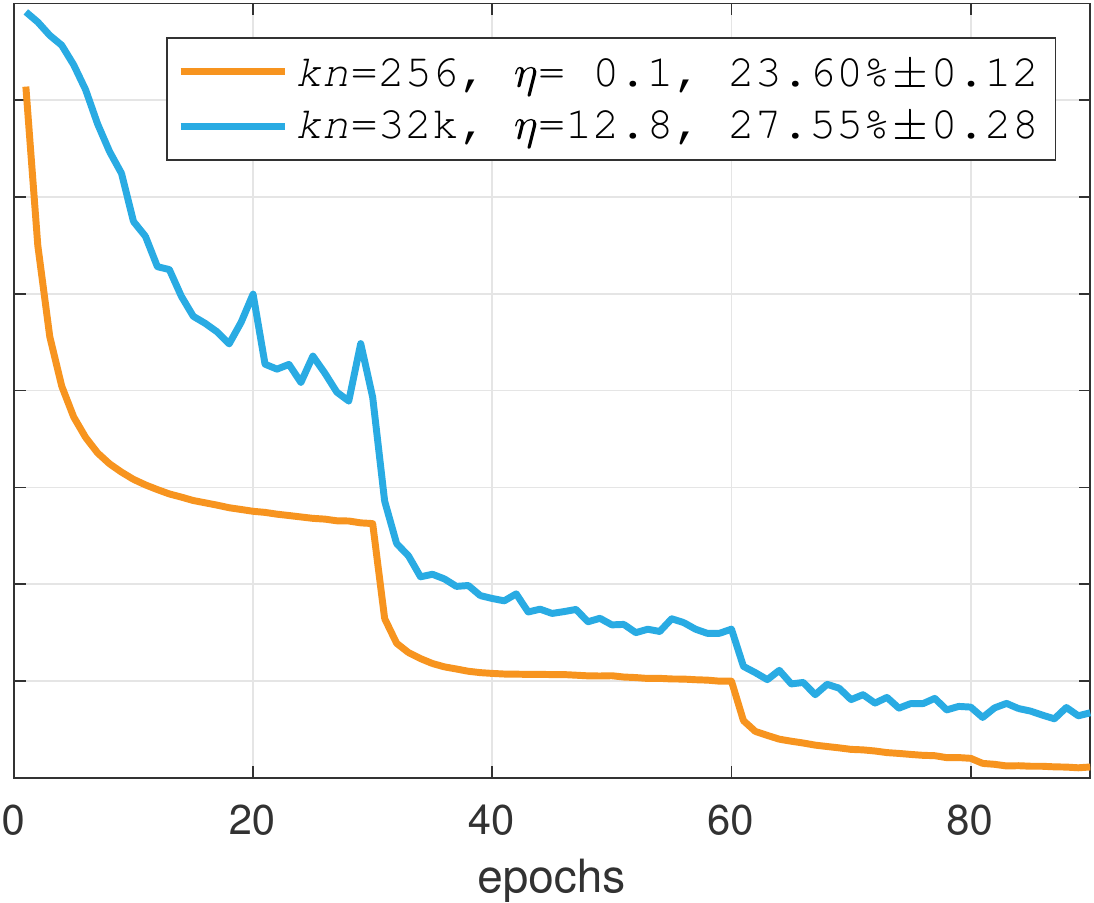}~
  \includegraphics[width=0.32\textwidth]{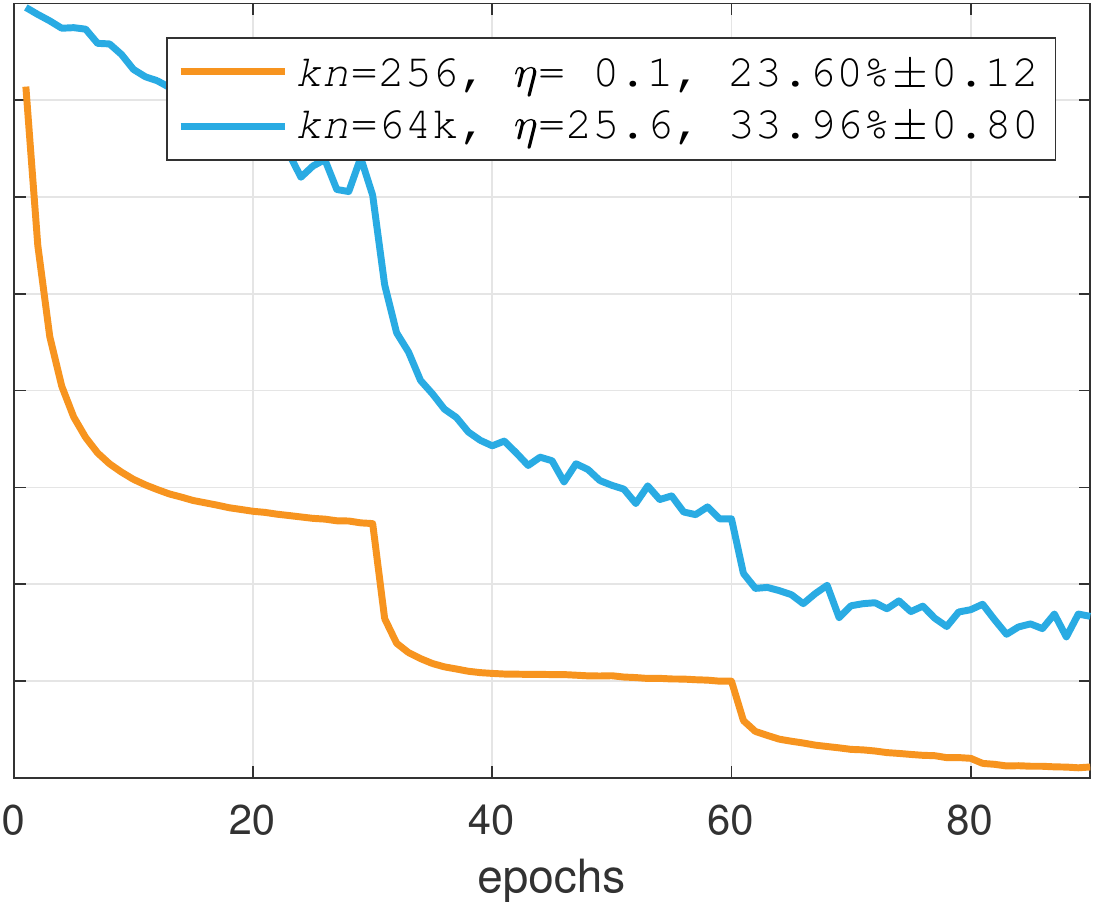}
  \caption{\bd{Training error \vs minibatch size.} Training error curves for the 256 minibatch baseline and larger minibatches using gradual warmup and the linear scaling rule. Note how the training curves closely match the baseline (aside from the warmup period) up through 8k minibatches. \emph{Validation} error (mean$\pm$std of 5 runs) is shown in the legend, along with minibatch size $kn$ and reference learning rate $\eta$.}
\label{fig:training:minibatch}
\end{figure*}

\begin{figure}[t]\centering
 \includegraphics[width=0.42\textwidth]{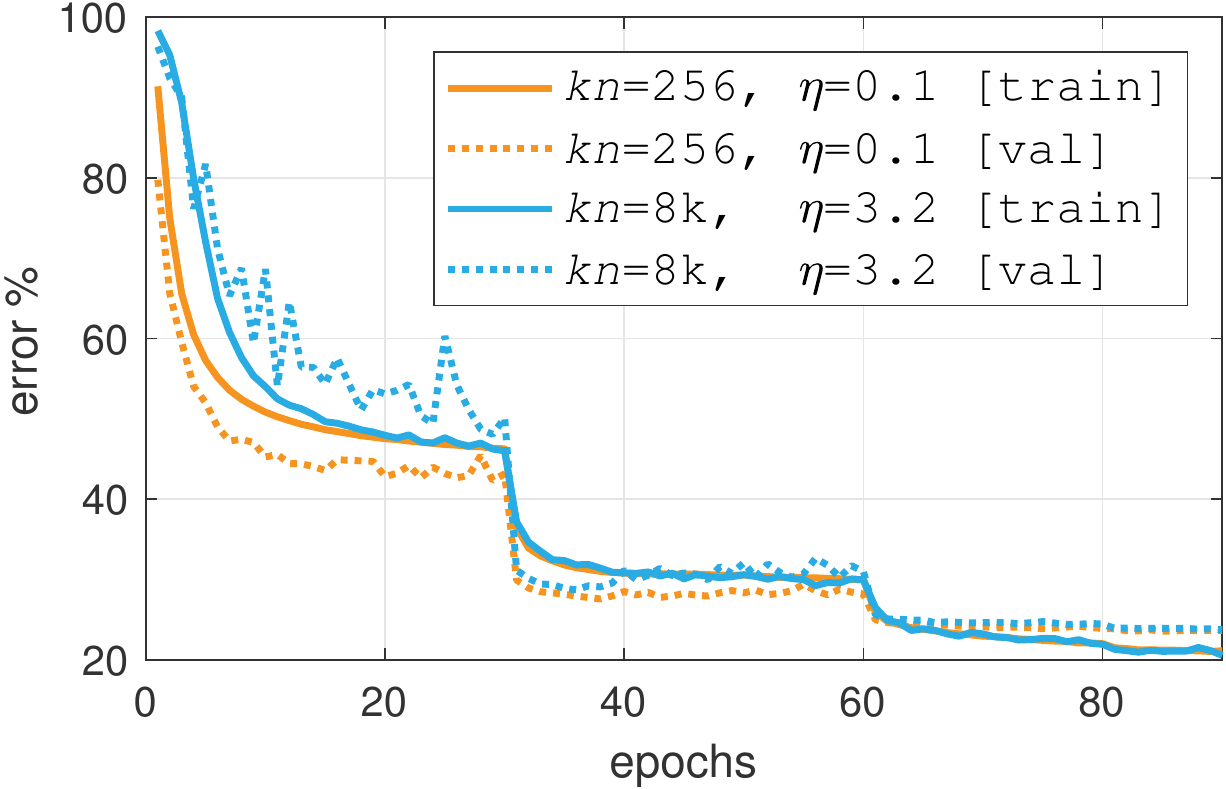}
\caption{\bd{Training and validation curves} for large minibatch SGD with gradual warmup \vs small minibatch SGD. Both sets of curves match closely after training for sufficient epochs. We note that the BN statistics (for inference only) are computed using \emph{running} average, which is updated less frequently with a large minibatch and thus is noisier in early training (this explains the larger variation of the validation error in early epochs).}
\label{fig:val}
\end{figure}

Finally, Figure \ref{fig:val} shows both the training and validation curves for the large minibatch training with gradual warmup. As can be seen, validation error starts to match the baseline closely after the second learning rate drop; actually, the validation curves can match earlier if BN statistics are recomputed prior to evaluating the error instead of using the running average (see also caption in Figure \ref{fig:val}).

\subsection{Analysis Experiments}\label{sec:exp:analysis}

\paragraph{Minibatch size \vs error.} Figure~\ref{fig:batchsize} (page 1) shows top-1 validation error for models trained with minibatch sizes ranging from of 64 to 65536 (64k). For all models we used the linear scaling rule and set the reference learning rate as $\eta=0.1\cdot\frac{kn}{256}$. For models with $kn>256$, we used the gradual warmup strategy always starting with $\eta=0.1$ and increasing linearly to the reference learning rate after 5 epochs. Figure~\ref{fig:batchsize} illustrates that validation error remains stable across a broad range of minibatch sizes, from 64 to 8k, after which it begins to increase. Beyond 64k training diverges when using the linear learning rate scaling rule.\footnote{We note that because of the availability of hardware, we \emph{simulated} distributed training of very large minibatches ($\ge$12k) on a single server by using multiple gradient accumulation steps between SGD updates. We have thoroughly verified that gradient accumulation on a single server yields equivalent results relative to distributed training.}

\paragraph{Training curves for various minibatch sizes.} Each of the nine plots in Figure~\ref{fig:training:minibatch} shows the top-1 training error curve for the 256 minibatch baseline (orange) and a second curve corresponding to different size minibatch (blue). Validation errors are shown in the plot legends. As minibatch size increases, all training curves show some divergence from the baseline at the start of training. However, in the cases where the final validation error closely matches the baseline ($kn\le8$k), the training curves also closely match after the initial epochs. When the validation errors do not match ($kn\ge16$k), there is a noticeable gap in the training curves for all epochs. This suggests that when comparing a new setting, the training curves can be used as a reliable proxy for success well before training finishes.

\begin{figure}[t]\centering
 \includegraphics[width=0.42\textwidth]{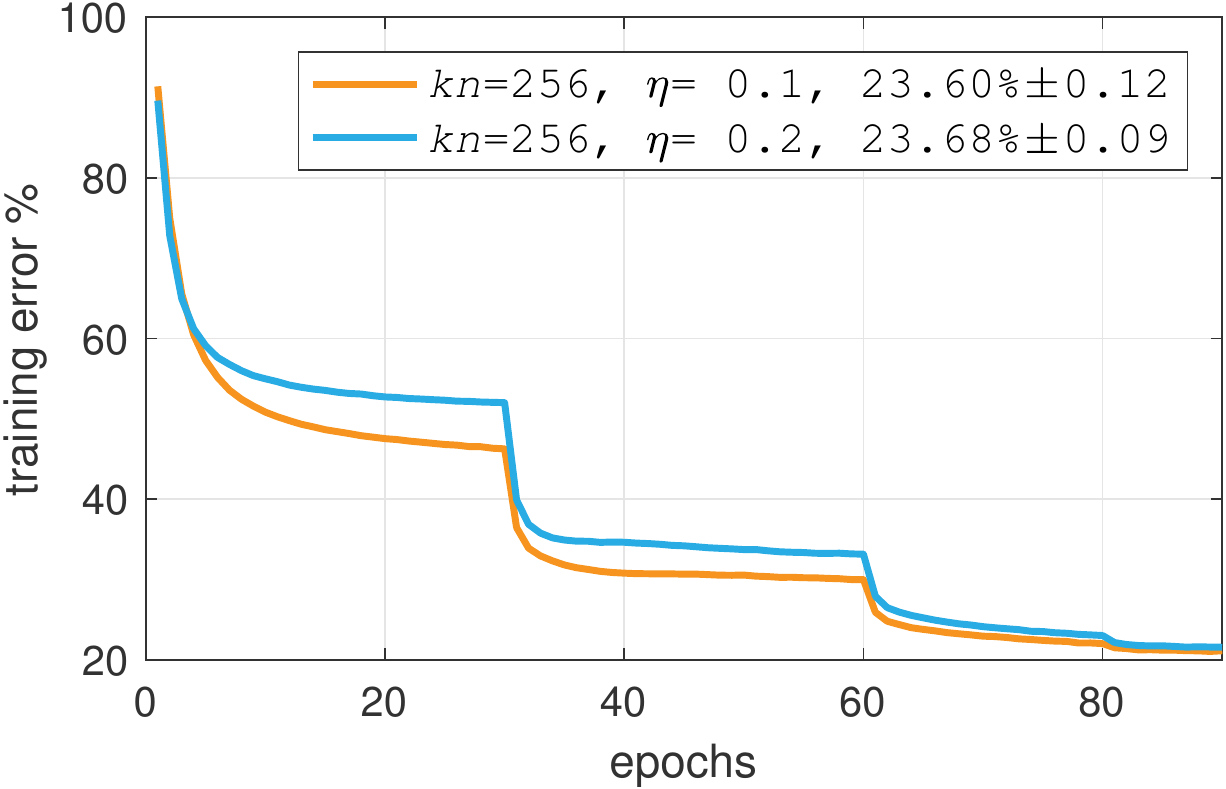}
\caption{\bd{Training curves for small minibatches with different learning rates $\eta$.} As expected, changing $\eta$ results in curves that \emph{do not match}. This is in contrast to changing batch-size (and linearly scaling $\eta$), which results in curves that \emph{do match}, \eg see Figure \ref{fig:training:minibatch}.}
\label{fig:lr}
\end{figure}

\paragraph{Alternative learning rate rules.} Table~\ref{tab:ablations:lr_scaling} shows results for multiple learning rates. For small minibatches ($kn=256$), $\eta=0.1$ gives best error but slightly smaller or larger $\eta$ also work well. When applying the linear scaling rule with a minibatch of 8k images, the optimum error is also achieved with $\eta=0.1 \cdot 32$, showing the successful application of the linear scaling rule. However, in this case results are more sensitive to changing $\eta$. In practice we suggest to use a minibatch size that is not close to the breaking point.

Figure~\ref{fig:lr} shows the training curves of a 256 minibatch using $\eta=0.1$ or $0.2$. It shows that changing the learning rate $\eta$ in general changes the overall shapes of the training curves, even if the final error is similar. Contrasting this result with the success of the linear scaling rule (that can match both the final error and the training curves when minibatch sizes change) may reveal some underlying invariance maintained between small and large minibatches.

We also show two alternative strategies: keeping $\eta$ fixed at 0.1 or using $0.1 \cdot \sqrt{32}$ according to the square root scaling rule that was justified theoretically in \cite{Krizhevsky2014} on grounds that it scales $\eta$ by the inverse amount of the reduction in the gradient estimator's standard deviation. For fair comparisons we also use gradual warmup for $0.1 \cdot \sqrt{32}$. Both policies work poorly in practice as the results show.

\begin{table}[t]\centering
\subfloat[\bd{Comparison of learning rate scaling rules.} A reference learning rate of $\eta=0.1$ works best for $kn=256$ (23.68\% error). The linear scaling rule suggests $\eta=0.1\cdot32$ when $kn=8$k, which again gives best performance (23.74\% error). Other ways of scaling $\eta$ give worse results.\label{tab:ablations:lr_scaling}]{
\makebox[1\linewidth]{
\tablestyle{12pt}{1.05}\begin{tabular}{ccc}
 $kn$ & $\eta$ & top-1 error (\%)\\
 \shline
 256 & $0.05$ & 23.92 $\pm$0.10 \\
 256 & $0.10$ & 23.60 $\pm$0.12 \\
 256 & $0.20$ & 23.68 $\pm$0.09 \\
 \hline
 8k  & $0.05 \cdot 32$ & 24.27 $\pm$0.08 \\
 8k  & $0.10 \cdot 32$ & 23.74 $\pm$0.09 \\
 8k  & $0.20 \cdot 32$ & 24.05 $\pm$0.18 \\
 \hline
 8k  & $0.10$ & 41.67 $\pm$0.10 \\
 8k  & $0.10 \cdot \sqrt{32}$ & 26.22 $\pm$0.03 \\
\end{tabular}}}\\
\subfloat[\bd{Batch normalization $\gamma$ initialization.} Initializing $\gamma=0$ in the \emph{last} BN layer of each residual block improves results for both small and large minibatches. This initialization leads to better optimization behavior which has a larger positive impact when training with large minibatches.\label{tab:ablations:bn_init}]{
\makebox[1\linewidth]{
\tablestyle{12pt}{1.05}\begin{tabular}{cccc}
 $kn$ & $\eta$ & $\gamma$-init & top-1 error (\%)\\
 \shline
 256  & 0.1 & 1.0 & 23.84 $\pm$0.18 \\
 256  & 0.1 & 0.0 & 23.60 $\pm$0.12 \\
 \hline
 8k   & 3.2 & 1.0 & 24.11 $\pm$0.07 \\
 8k   & 3.2 & 0.0 & 23.74 $\pm$0.09 \\
\end{tabular}}}\\
\subfloat[\bd{The linear scaling rule applied to ResNet-101.} The difference in error is about 0.3\% between small and large minibatch training.\label{tab:ablations:resnet101}]{
\makebox[1\linewidth]{
\tablestyle{12pt}{1.05}\begin{tabular}{c|ccc}
model type & $kn$ & $\eta$ & top-1 error (\%)\\
 \shline
 ResNet-101 & 256 & 0.1 & 22.08 $\pm$0.06 \\
 ResNet-101 & 8k  & 3.2 & 22.36 $\pm$0.09 \\
\end{tabular}}}\vspace{3mm}
\caption{\textbf{ImageNet classification experiments.} Unless noted all experiments use ResNet-50 and are averaged over 5 trials.}
\label{tab:ablations}
\end{table}

\paragraph{Batch Normalization $\gamma$ initialization.} Table~\ref{tab:ablations:bn_init} controls for the impact of the new BN $\gamma$ initialization introduced in \S\ref{sec:exp:settings}. We show results for minibatch sizes 256 and 8k with the standard BN initialization ($\gamma=1$ for all BN layers) and with our initialization ($\gamma=0$ for the final BN layer of each residual block). The results show improved performance with $\gamma=0$ for both minibatch sizes, and the improvement is slightly larger for the 8k minibatch size. This behavior also suggests that large minibatches are more easily affected by optimization difficulties. We expect that improved optimization and initialization methods will help push the boundary of large minibatch training.

\paragraph{ResNet-101.} Results for ResNet-101 \cite{He2016} are shown in Table~\ref{tab:ablations:resnet101}. Training ResNet-101 with a batch-size of $kn=8$k and a linearly scaled $\eta=3.2$ results in an error of 22.36\% \vs the $kn=256$ baseline which achieves 22.08\% with $\eta=0.1$. In other words, ResNet-101 trained with minibatch 8k has a small 0.28\% increase in error \vs the baseline. It is likely that the minibatch size of 8k lies on the edge of the useful minibatch training regime for ResNet-101, similarly to ResNet-50 (see Figure \ref{fig:batchsize}).

The training time of ResNet-101 is 92.5 minutes in our implementation using 256 Tesla P100 GPUs and a minibatch size of 8k. We believe this is a compelling result if the speed-accuracy tradeoff of ResNet-101 is preferred.

\paragraph{ImageNet-5k.} Observing the sharp increase in validation error between minibatch sizes of 8k and 16k on ImageNet-1k (Figure \ref{fig:batchsize}), a natural question is if the position of this `elbow' in the error curve is a function of dataset information content. To investigate this question, we adopt the ImageNet-5k dataset suggested by Xie \etal \cite{xie2017} that extends ImageNet-1k to 6.8 million images (roughly 5$\x$ larger) by adding 4k additional categories from ImageNet-22k \cite{Russakovsky2015}. We evaluate the 1k-way classification error on the original ImageNet-1k validation set as in \cite{xie2017}.

\begin{figure}[t]\centering
\includegraphics[width=.99\linewidth]{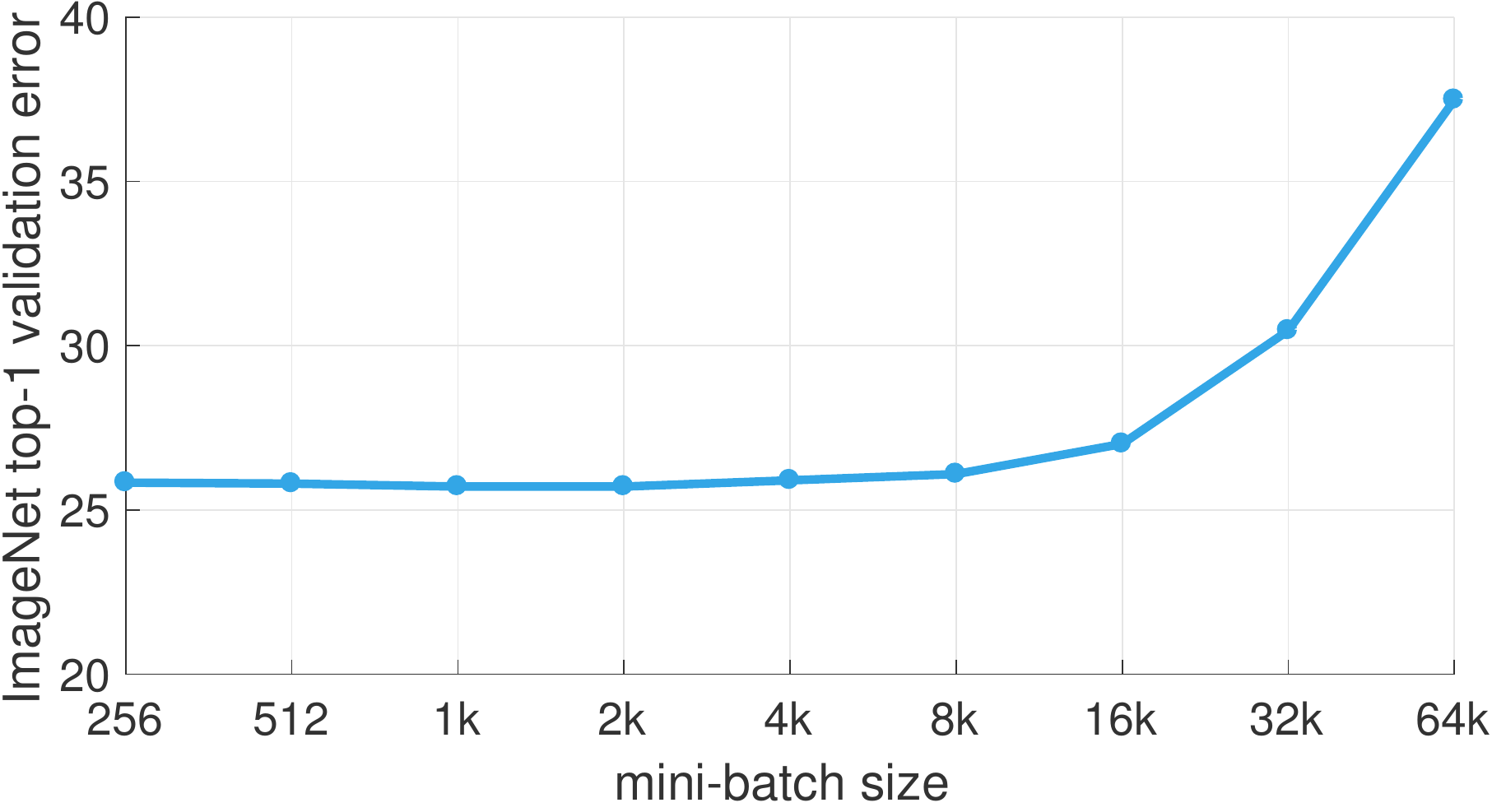}
\caption{\bd{ImageNet-5k} top-1 validation error \vs minibatch size with a fixed 90 epoch training schedule. The curve is qualitatively similar to results on ImageNet-1k (Figure \ref{fig:batchsize}) showing that a 5$\x$ increase in training data does not lead to a significant change in the maximum effective minibatch size.}
\label{fig:batchsizeIN5k}\vspace{2mm}
\end{figure}

The minibatch size \vs validation error curve for ImageNet-5k is shown in Figure \ref{fig:batchsizeIN5k}. Qualitatively, the curve is very similar to the ImageNet-1k curve, showing that for practitioners it is unlikely that even a 5$\x$ increase in dataset size will automatically lead to a meaningful increase in useable minibatch size. Quantitatively, using an 8k minibatch increases the validation error by 0.26\% from 25.83\% for a 256 minibatch to 26.09\%. An understanding of the precise relationship between generalization error, minibatch size, and dataset information content is open for future work.

\begin{table}[t]\centering
\subfloat[\bd{Transfer learning of large minibatch pre-training to Mask R-CNN.} Box and mask AP (on COCO \symb{minival}) are nearly identical for ResNet-50 models pre-trained with minibatches from 256 to 8k examples. With a minibatch pre-training size of 16k both ImageNet validation error \emph{and} COCO AP deteriorate. This indicates that as long as ImageNet error is matched, large minibatches do not degrade transfer learning performance.\label{tab:det:pretraining}]{
\makebox[1\linewidth]{
\tablestyle{8pt}{1.05}\begin{tabular}{ccc|cc}
\multicolumn{3}{c|}{ImageNet pre-training} & \multicolumn{2}{c}{COCO} \\
$kn$ & $\eta$ & top-1 error (\%) & box AP (\%) & mask AP (\%)\\[.1em]
 \shline
 256 & 0.1 & 23.60 $\pm$0.12 & 35.9 $\pm$0.1 & 33.9 $\pm$0.1 \\
 512 & 0.2 & 23.48 $\pm$0.09 & 35.8 $\pm$0.1 & 33.8 $\pm$0.2 \\
 1k  & 0.4 & 23.53 $\pm$0.08 & 35.9 $\pm$0.2 & 33.9 $\pm$0.2 \\
 2k  & 0.8 & 23.49 $\pm$0.11 & 35.9 $\pm$0.1 & 33.9 $\pm$0.1 \\
 4k  & 1.6 & 23.56 $\pm$0.12 & 35.8 $\pm$0.1 & 33.8 $\pm$0.1 \\
 8k  & 3.2 & 23.74 $\pm$0.09 & 35.8 $\pm$0.1 & 33.9 $\pm$0.2 \\
 \hline
 \demph{16k} & \demph{6.4} &\demph{24.79 $\pm$0.27} & \demph{35.1 $\pm$0.3} & \demph{33.2 $\pm$0.3} \\
\end{tabular}}}\\
\subfloat[\bd{Linear learning rate scaling applied to Mask R-CNN.} Using the single ResNet-50 model from \cite{He2016} (thus no std is reported), we train Mask R-CNN using using from 1 to 8 GPUs following the linear learning rate scaling rule. Box and mask AP are nearly identical across all configurations showing the successful generalization of the rule beyond classification.\label{tab:det:lr_scaling}]{
\makebox[1\linewidth]{
\tablestyle{4pt}{1.05}\begin{tabular}{c|ccccc}
\# GPUs & $kn$ & $\eta\cdot1000$ & iterations & box AP (\%) & mask AP (\%) \\[.1em]
 \shline
 1 & 2  & 2.5  & 1,280,000 & 35.7 & 33.6 \\
 2 & 4  & 5.0  & 640,000   & 35.7 & 33.7 \\
 4 & 8  & 10.0 & 320,000   & 35.7 & 33.5 \\
 8 & 16 & 20.0 & 160,000   & 35.6 & 33.6 \\
\end{tabular}}}
\vspace{2mm}
\caption{\textbf{Object detection on COCO with Mask R-CNN \cite{He2017}.}}
\label{tab:det}\vspace{-2mm}
\end{table}

\subsection{Generalization to Detection and Segmentation}\label{sec:exp:detection}

A low error rate on ImageNet is not typically an end goal. Instead, the utility of ImageNet training lies in learning good features that transfer, or generalize well, to related tasks. A question of key importance is if the features learned with large minibatches generalize as well as the features learned with small minibatches?

To test this, we adopt the object detection and instance segmentation tasks on COCO \cite{Lin2014} as these advanced perception tasks benefit substantially from ImageNet pre-training \cite{Girshick2014}. We use the recently developed Mask R-CNN \cite{He2017} system that is capable of learning to detect and segment object instances. We follow all of the hyper-parameter settings used in \cite{He2017} and only change the ResNet-50 model used to initialize Mask R-CNN training. We train Mask R-CNN on the COCO \symb{trainval35k} split and report results on the 5k image \symb{minival} split used in \cite{He2017}.

It is interesting to note that the concept of minibatch size in Mask R-CNN is different from the classification setting. As an extension of the \emph{image-centric} Fast/Faster R-CNN \cite{Girshick2015,Ren2015}, Mask R-CNN exhibits \emph{different minibatch sizes for different layers}: the network backbone uses two images (per GPU), but each image contributes 512 Regions-of-Interest for computing classification (multinomial cross-entropy), bounding-box regression (smooth-L1/Huber), and pixel-wise mask ($28\x28$ binomial cross-entropy) losses. This diverse set of minibatch sizes and loss functions provides a good test case to the robustness of our approach.

\begin{figure}[t]\centering
\includegraphics[width=.99\linewidth]{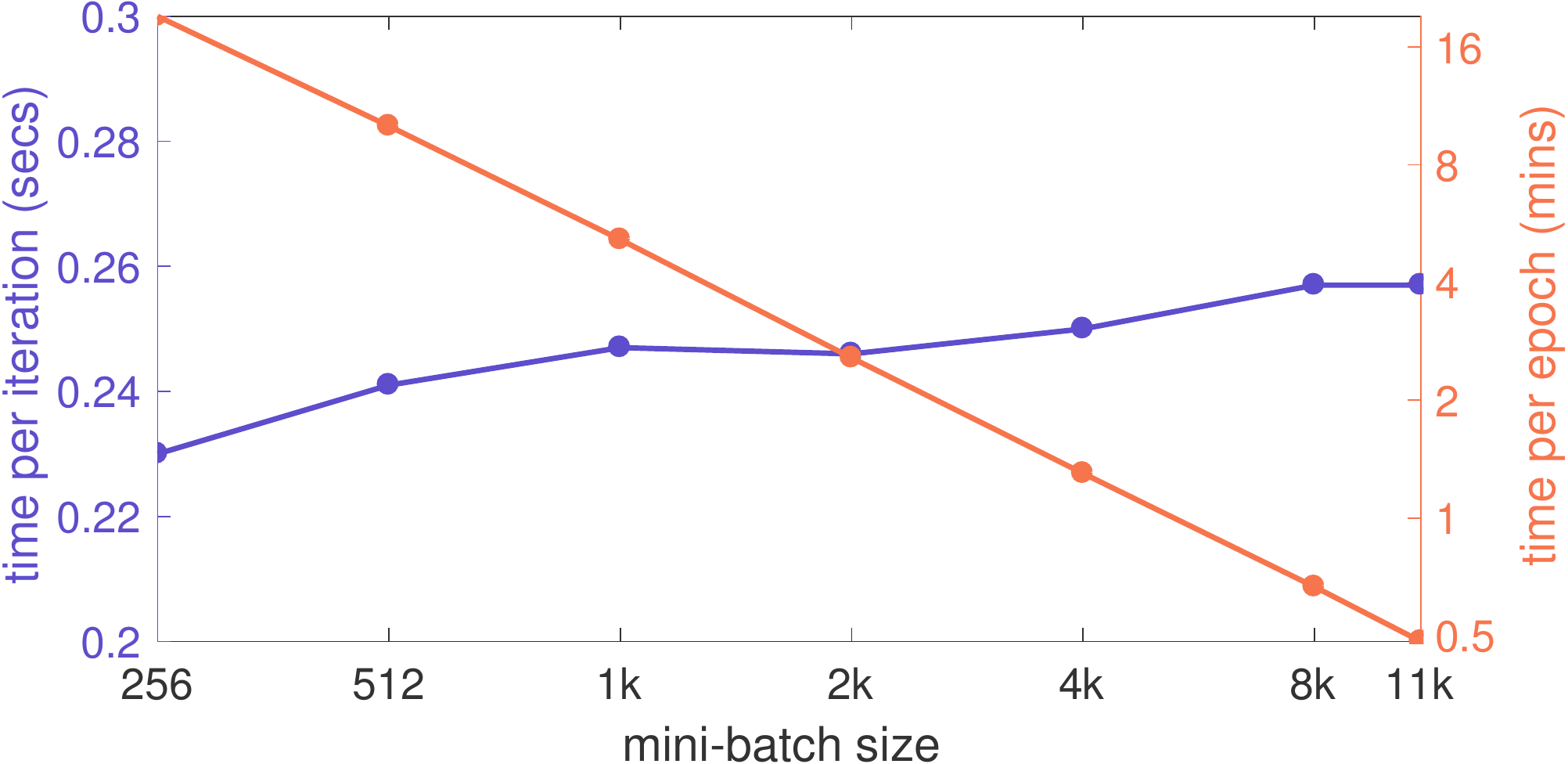}
\caption{\bd{Distributed synchronous SGD timing.} Time per iteration (seconds) and time per ImageNet epoch (minutes) for training with different minibatch sizes. The baseline ($kn=256$) uses 8 GPUs in a single server , while all other training runs distribute training over ($kn/256$) server. With 352 GPUs (44 servers) our implementation completes one pass over all $\app$1.28 million ImageNet training images in about 30 seconds.}
\label{fig:time}
\end{figure}

\paragraph{Transfer learning from large minibatch pre-training.} To test how large minibatch \emph{pre-training} effects Mask R-CNN, we take ResNet-50 models trained on ImageNet-1k with 256 to 16k minibatches and use them to initialize Mask R-CNN training. For each minibatch size we pre-train 5 models and then train Mask R-CNN using all 5 models on COCO (35 models total). We report the mean box and mask APs, averaged over the 5 trials, in Table~\ref{tab:det:pretraining}. The results show that as long as ImageNet validation error is kept low, which is true up to 8k batch size, generalization to object detection matches the AP of the small minibatch baseline. We emphasize that we observed \emph{no generalization issues} when transferring across datasets (from ImageNet to COCO) and across tasks (from classification to detection/segmentation) using models trained with large minibatches.

\paragraph{Linear scaling rule applied to Mask R-CNN.} We also show evidence of the generality of the linear scaling rule using Mask R-CNN. In fact, this rule was already used without explicit discussion in \cite{He2016} and was applied effectively as the default Mask R-CNN training scheme when using 8 GPUs. Table~\ref{tab:det:lr_scaling} provides experimental results showing that when training with 1, 2, 4, or 8 GPUs the linear learning rate rule results in constant box and mask AP. For these experiments, we initialize Mask R-CNN from the released MSRA ResNet-50 model, as was done in \cite{He2017}.

\subsection{Run Time}\label{sec:exp:timing}

\begin{figure}[t]\centering
\includegraphics[width=.99\linewidth]{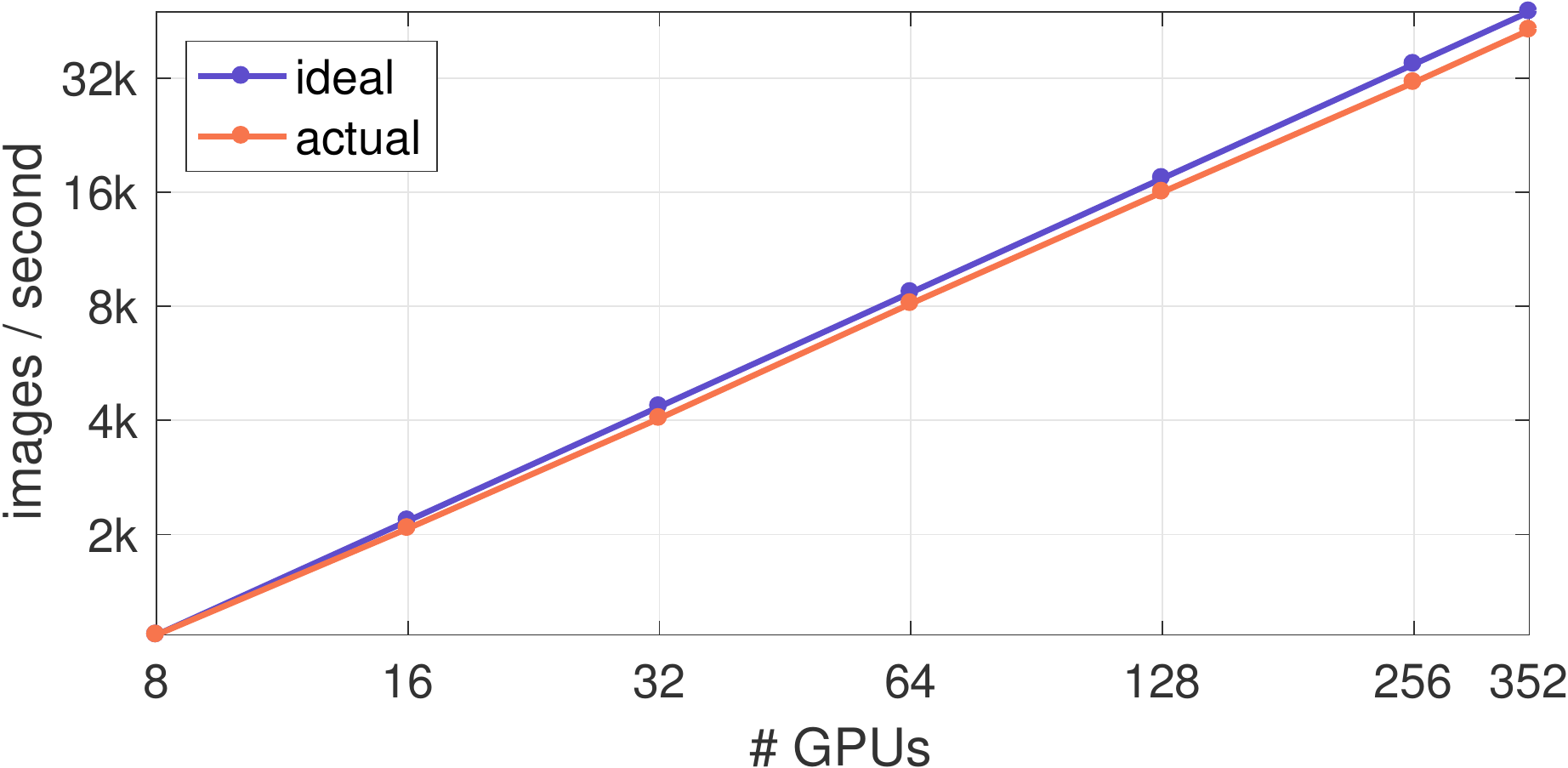}
\caption{\bd{Distributed synchronous SGD throughput.} The small overhead when moving from a single server with 8 GPUs to multi-server distributed training (Figure~\ref{fig:time}, blue curve) results in linear throughput scaling that is marginally below ideal scaling (\app 90\% efficiency). Most of the allreduce communication time is hidden by pipelining allreduce operations with gradient computation. Moreover, this is achieved with commodity Ethernet hardware.}
\label{fig:throughput}
\end{figure}

Figure \ref{fig:time} shows two visualizations of the run time characteristics of our system. The blue curve is the time per iteration as minibatch size varies from 256 to 11264 (11k). Notably this curve is relatively flat and the time per iteration increases only 12\% while scaling the minibatch size by 44$\x$. Visualized another way, the orange curve shows the approximately linear decrease in time per epoch from over 16 minutes to just 30 seconds. Run time performance can also be viewed in terms of throughput (images / second), as shown in Figure \ref{fig:throughput}. Relative to a perfectly efficient extrapolation of the 8 GPU baseline, our implementation achieves \app90\% scaling efficiency.

{\small\paragraph{Acknowledgements.} We would like to thank Leon Bottou for helpful discussions on theoretical background, Jerry Pan and Christian Puhrsch for discussions on efficient data loading, Andrew Dye for help with debugging distributed training, and Kevin Lee, Brian Dodds, Jia Ning, Koh Yew Thoon, Micah Harris, and John Volk for Big Basin and hardware support.}

{\linespread{.95}\small\bibliographystyle{ieee}\bibliography{imagenet.bib}}

\end{document}